\documentclass[runningheads]{llncs}
\usepackage{graphicx}
\usepackage{subfiles} 

\usepackage{tikz}
\usepackage{comment}
\usepackage{amsmath,amssymb} 
\usepackage{color}
\usepackage{multirow}
\usepackage{float}
\restylefloat{table}

\usepackage[accsupp]{axessibility}  
\usepackage{hyperref}

\begin{document}
\pagestyle{headings}
\mainmatter
\def\ECCVSubNumber{13}  
\def\etal{\emph{et al}.}

\title{Visible-Infrared Person Re-Identification Using Privileged Intermediate Information} 

\titlerunning{Visible-Infrared Person-ReID Using LUPI}
%
\author{
    Mahdi Alehdaghi\orcidID{0000-0001-6258-8362} \and
    Arthur Josi\orcidID{0000-0002-4362-4898} \and
    Rafael M. O. Cruz\orcidID{0000-0001-9446-1040} \and
    Eric Granger\orcidID{0000-0001-6116-7945}
} 
\authorrunning{M. Alehdaghi et al.}
%
\institute{Laboratoire d’imagerie, de vision et d’intelligence artificielle (LIVIA)\\ 
Dept. of Systems Engineering,  ETS Montreal, Canada\\
\email{\{mahdi.alehdaghi.1, arthur.josi.1\}@ens.etsmtl.ca, \\ \{rafael.menelau-cruz, eric.granger\}@etsmtl.ca}}

\maketitle

\begin{abstract}
Visible-infrared person re-identification (ReID) aims to recognize a same person of interest across a network of RGB and IR cameras. Some deep learning (DL) models have directly incorporated both modalities to discriminate persons in a joint representation space. However, this cross-modal ReID problem remains challenging due to the large domain shift in data distributions between RGB and IR modalities. 
This paper introduces a novel approach for a creating intermediate virtual domain that acts as bridges between the two main domains (i.e., RGB and IR modalities) during training. This intermediate domain is considered as privileged information (PI) that is unavailable at test time, and allows formulating this cross-modal matching task as a problem in learning under privileged information (LUPI). We devised a new method to generate images between visible and infrared domains that provide  additional information to train a deep ReID model through an intermediate domain adaptation. In particular, by employing color-free and multi-step triplet loss objectives during training, our method provides common feature representation spaces  that are robust to large visible-infrared domain shifts. 
Experimental results on challenging visible-infrared ReID datasets indicate that our proposed approach consistently improves matching accuracy, without any computational overhead at test time. 
The code is available at: \href{https://github.com/alehdaghi/Cross-Modal-Re-ID-via-LUPI}{https://github.com/alehdaghi/Cross-Modal-Re-ID-via-LUPI}
\end{abstract}

\section{Introduction}
\label{sec:intro}

Person re-identification (ReID) has a growing interest in several real-world computer vision applications, such as video monitoring and surveillance, search and retrieval, and pedestrian tracking for autonomous driving. It involves matching images or videos to retrieve individuals captured across a  distributed network of non-overlapping video cameras. The non-rigid structure of human bodies, variations in capture conditions in the wild, due to changes in illumination, motion blur, resolution, pose, view point, and occlusion and background clutter leads to high intra-person and low inter-person variability~\cite{kiran2021holistic,mekhazni2020unsupervised}. Despite these considerable challenges, recent progress in deep learning (DL) has allowed to develop state-of-art person ReID models that can achieve a high  level of accuracy when trained on large fully-labeled image datasets. However, the domain shift typically associated with diverse operational capture conditions (e.g., camera viewpoints and lighting) may translate to a significant decline in performance. Most image-based ReID methods are formulated as a single-modality retrieval problem, where features extracted from input query images are matched against reference gallery images. For example, where all the images are captured by visible cameras, encouraging results have been observed \cite{jia2020similarity}. Visible cameras, however, are somewhat limited in real-world applications because of insufficient discrimination under poor lighting conditions, e.g., in night time surveillance.

Most modern surveillance systems operate in dual modes, i.e., in the visible mode during the day and switch to the infrared mode at night. In this way, a supplementary task is also developed alongside visual ReID: Given a visible target image, the goal is to match it with the infrared image of the corresponding individual. This cross-modal image matching task is named Visible-Infrared person re-identification (VI-ReID). Alongside Intra-class variations (e.g., viewpoint, pose, illumination, and background clutter) and noisy samples (e.g., misalignment and occlusion), the cross-modal discrepancies between visible and infrared images make VI-ReID extremely challenging.

VI-ReID currently comprises two components: feature extraction and distance measurement. An imperative objective of feature extraction is to extract discriminating features to separate different individuals captured by several cameras. For example, in \cite{jiang2020crossGranularity,EDFL:journals/corr/abs-1907-09659,regDB,SYSU,wu2017rgb} two separated CNN are used as backbone for feature extraction for infrared and visible images. Based on the extracted features, distance measurement seeks to measure the similarity between two images and then optimize the feature extractor backbones, aiming to increase the similarity for pairs with the same individual while decreasing it otherwise. Considering the cross-modal task, the model needs to leverage and extract the shared information between two modalities. Such information should be modality invariant and also can be disentangled from modality-specific ones, for example, Yang et al. \cite{paired-images2}, Choi et al. \cite{HI-CMD}, and Kniaz et al. \cite{kniaz2018thermalgan} used GANs models to translate visible and infrared images. Since there are significant differences in the visual attributes and statistics of visible/infrared images \cite{wu2017rgb}, one special challenge lies in bridging the modality gap between the visible and infrared images. Multiple approaches \cite{Fan2020CrossSpectrumDP,DZP,HAT} focus on spectrum mining between visual image colors and infrared and make gray images the bridge. These approaches consider the bridge as one static domain between infrared and visible. For example, the method proposed in \cite{HAT} used gray images as the middle modality, and \cite{ye2021channel} used augmentation, selecting one channel randomly from the color spectrum of optic inputs. In fact, making proper intermediate images for tackling the large gap between infrared and visible is crucial for the model training process. But, using gray scale only or one channel from colored images is not perfectly able to link between colored and infrared images. For example, a person with same colored clothes, may have different infrared view in different environment. So, it motivates us to use more effective way for such intermediate domains.  

In this paper, an effective way is proposed for bridging RGB and IR images by selecting different amounts of the visible spectrum. The learning strategy is based on the learning under privileged information (LUPI) paradigm \cite{LUPI}, where image data from an intermediate domain is generated for training DL models for cross-modal person ReID. The intermediate domain allow the model to extract a common yet discriminant feature representation for visible and infrared modalities, and reduce the domain shift between the distribution of these  modalities (as analysed section \ref{sec:ablation_study}).  Besides LUPI, our approach is also inspired by methods in domain adaptation (DA), where a model (trained on source data) is adapted to a target domain by learning a common source-target representation space. Given a large domain shift, multi-task DA methods adapt models through one or more intermediate domains \cite{deepDA_survey}. 

Our training model is comprised of 2 CNN backbones -- two students for the main modalities, and a teacher sub-model for the intermediate modality. Instead of learning the infrared model from visible images (and vice versa), we generated intermediate images from visible inputs with less visual difference from infrared inputs, while preserving discriminating information from visible images. We also introduce a simple but effective spatial image augmentation mechanism intended to force the main branches of our model to focus on learning from the PI branch and consequently avoid focusing on modality-specific information. We show that common semantic information between visible and infrared images is, to a greater extent, dependent on textures. Therefore, intermediate images with the same texture pattern as visible images that do not contain color information may be a suitable intermediate step in linking to RGB and IR. During testing, our proposed model uses only one of the branches for the cross-modal ReID, so the generated intermediate domain does not impact the inference time. During training, a one-channel image is created from visual inputs with random color combinations. There are two reasons for this. First, they are not dependent on color information, making the feature representation of visual inputs less color discriminative. Second, the visual difference between infrared and grayscale images is smaller than between color and such color-free images. 

The main contributions of this paper are summarized as follows.
     (1) The cross-modal VI-ReID problem is reformulated according to the LUPI paradigm. To address the large domain shifts between RGB and IR data distributions, we propose leveraging related PI as intermediate domains to train the CNN backbone.
     (2) An effective method is introduced to create intermediate virtual domains -- the PI -- from visible images based on a random linear combination of color channels. This allows bridging the RGB and IR modality domains during training.
     (3) Multi-step triplet and color-free losses are proposed to learn common feature representations for visible and infrared images, and also provide color-independent intermediate images as an effective bridge between modalities. In addition, the cross-modal distance between  modalities is minimized according to intermediate features. 
    (4) An extensive set of experiments on the challenging SYSU-MM01 \cite{SYSU} and RegDB \cite{regDB} datasets show that our proposed approach can significantly outperform state-of-art methods for cross-modal visible-infrared person ReID, yet incur no computational overhead at test time. 

\section{Related Work}

\noindent \textbf{Cross-Modality Person Re-ID:}
Visible-Infrared Person ReID (VI-ReID) focuses on matching visible daytime images against infrared nighttime images of a person for cross-modal identification problems \cite{chen2021neural,cm-gan,fu2021cm,hao2020modality,park2021learning,xu2021cross,HCML,hetero-center}. A general view of pipelines to train DL models for cross-modal matching is categorized as \cite{cross-survey}:
(1) Representation: How to represent and summarize multimodal data to better exploit the modality-invariant and complementarity or even redundancy of multiple modalities.  
(2) Translation: How to map data from one modality to another, which could be heterogeneous, and the relationship between them is often open-ended or subjective.  
(3) Co-learning: How to transfer knowledge between modalities, their representation, and their predictive models. 

A challenging problems encountered in cross-modal matching is the large discrepancies between visible and infrared modalities. Wu Ancong \etal \cite{DZP}, for example, attempts to reduce the discrepancies in the spectrum level between color channels and infrared values by using a grayscale version of colored images.  Fan \etal \cite{Fan2020CrossSpectrumDP} used an image created from a combination of infrared, visible and Gray channels information, as using gray only may result in the omission of specific information in the visible spectrum.  \cite{liu2021sfanet,HAT} used the grayscale modality as a bridge between the infrared and visible ones in a triple CNN model training. \cite{HAT} used gray images, produced from the visible images, and constrained the last convolution layer of the Gray and visible backbones to produce similar features. Using gray alongside visible and infrared improves the performance \cite{liu2021sfanet,HAT}, but it is limited to a specific version of colored images. 
Part-based representation models learn to extract part/region aggregated features, making them robust against misalignment or occlusion. \cite{hetero-center} introduced horizontal-divided region features at the end of a deep model to learn and represent local sub-feature. To capture the relations across multiple body parts, Intra-modality Weighted-Part Aggregation was presented in \cite{DDAG} to mine the contextual information in local parts and formulate an enhanced part-aggregated representation. Finally, some models have been proposed to extract modality-invariant information from each modality, and then training their CNN backbone in loss functions in the feature space. In order to learn shared features across modalities, a dual-constrained top-ranking loss, and a two-stream network have been developed in \cite{Bi-Di_Center-Constrained}. Adversarial approaches to create a common feature space are used in \cite{cm-gan} and \cite{adv-modal}, where an embedding model tries to fool a modality discriminator model.

\noindent \textbf{Generative Methods for Cross-Modality ReID:}
ThermalGAN \cite{kniaz2018thermalgan} used an autoencoder to translate visible images to infrared in pixel-level similarity. AlignGAN \cite{Wang_2019_ICCV_AlignGAN} has two encoder-decoders to translate infrared to visible, and vice versa from shared encoded feature space, which is used for matching in the testing stage. 
\cite{HI-CMD,JSIA-paired-images1,paired-images2} contain two specific encoders for infrared and visible, as well as a shared encoder to separate the modality-invariant information from the modality-specific one. By using GAN-based methods, synthetic images are generated and are especially conditioned by a modality-invariant representation which works at diminishing the pixel-level modality discrepancy. However, these methods are complex, and generated images do not possess enough quality to bridge the modality gap between synthetic and target data. To avoid using GAN models for generation of intermediate images, our approach instead relies on spectrum information, making our approach much faster.

\noindent \textbf{Learning Under Privileged Information:}
Traditionally, supervised learning focuses on finding a decision function that minimizes the generalization error on on a labeled training data. In the case of easy learning problems and with larger training datasets, the function found by the learner typically converges rapidly to the optimal value. On the other hand, if the learning problem is challenging and the learner's space of decision functions is large, the convergence rate (or learning rate) will be slow \cite{pechyony2010theory}. The idea of LUPI was first proposed by Vapnik \& Vashist in \cite{vapnik2009new} to increase the convergence rate of Support Vector Machines (SVMs). Hoffman \etal \cite{hoffman2016learning} was the first to introduce network hallucination, aiming to introduce PI knowledge through CNN by using an additional CNN stream. This additional stream learns to extract depth-related features (PI) from the RGB modality thanks to knowledge distillation.

The LUPI approach had been applied with success in various domains, and its benefits go beyond faster convergence \cite{choi2017learning,kampffmeyer2018urban,kumar2021improved,lezama2017not,saputra2020deeptio}. Instead of hallucinating the modality through another network, Pande \cite{pande2019adversarial} learned a GAN to generate the PI-related features at inference time. Crasto et al. \cite{crasto2019mars} directly distilled the PI knowledge (optical flow) to the current RGB stream, learning with both the cross-entropy and MSE losses, and avoiding the extra inference parameters. \cite{cho2021dealing} recently extended the approach to Visual Question Answering (VQA) method, playing with multiple teachers. To the best of our knowledge, our work is the first application of the LUPI paradigm in the context of cross-modal person VI-ReID, where each modality is considered as a PI for the other. Moreover, we introduce knowledge from the PI through new losses and layers with shared parameters. It has the advantage of avoiding the extra parameters related to LUPI. Our innovative method is suitable for a wide range of cross-modal ReID frameworks 


\section{Proposed Method}

In cross-modality matching, the main learning objective is to allow for extraction of image features that are invariant across modalities. Let $\mathcal{V}=\{v_i\}_1^n$ be the visible and $\mathcal{T}=\{t_i\}_{n+1}^{n+m}$ be the infrared or thermal images of several persons. Also, $\mathcal{Y} = \{y_i\}_1^{n+m}$ contains the identity of each person in dataset. The multi-modal dataset is represented as $\mathcal{D} = \{ \mathcal{D}_{tr}, \mathcal{D}_{te}\}$, where $\mathcal{D}_{tr}$ denotes the training data and $\mathcal{D}_{te}$ denotes the testing data, in which $\mathcal{D}_{tr} = \{\mathcal{V}_{tr}, \mathcal{T}_{tr}\}$ and $\mathcal{D}_{te} = \{\mathcal{V}_{te}, \mathcal{T}_{te}\}$.   The learning objective is to optimize the relation between the extracted features $\mathbf{f}^v_i = \varphi(v_i;\Theta)$ of visible images, space and the features $\mathbf{f}^t_j = \varphi(t_j;\Theta)$ of near-infrared images, denoted by:
\begin{equation}
\label{eq:problem}
    \mathcal{L} = \sum{l(\mathbf{f}^v_i, \mathbf{f}^t_j, y_i, y_j)}
\end{equation}
where $y_i$ and $y_j$ are the annotated training labels for each image and $\mathbf{f}^v_i, \mathbf{f}^t_j \in \mathbb{R}^d$, $d$ being the features dimension.

\subsection{LUPI Framework}

With LUPI \cite{lambert2018deep}, the DL model has access to PI, $x^\star$, only during the training while main information $x$ is available during both training and inference phases. When using the cross-modal representative model, two modalities should be used to represent inputs during the training phase so that the models can learn to use the cross-modal information, whereas only individual inputs will be fed to the model during the inference phase. That is, the missing modality at inference can be seen as a PI.

\begin{figure}
\centering
\begin{tabular}{c@{\hspace{1cm}}c}
\centering
{\includegraphics[width=5.3cm]{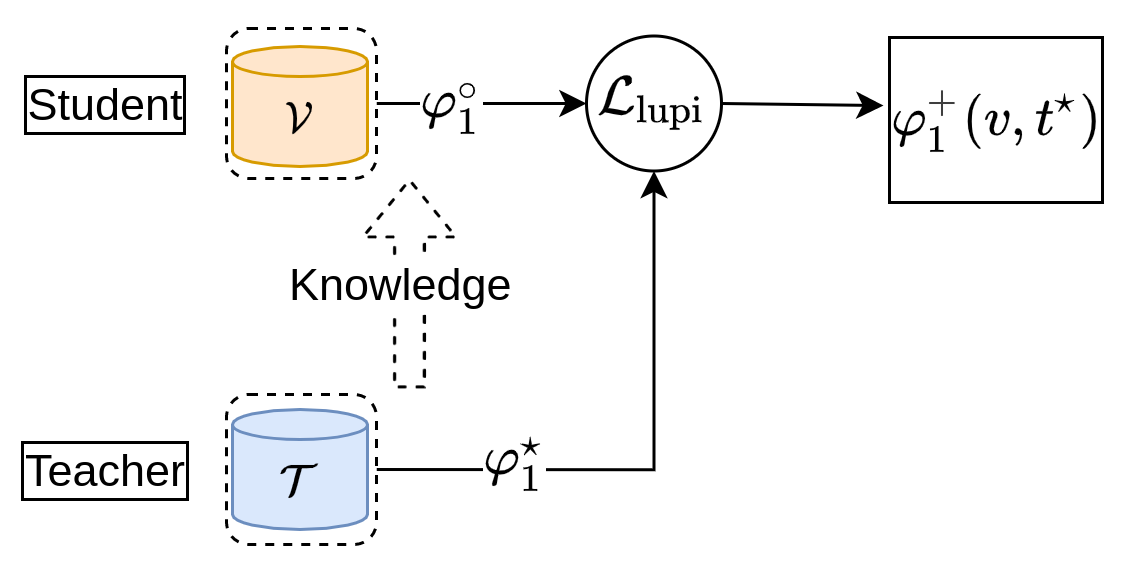}} & \multirow{3}{*}[1.5cm]{{\includegraphics[width=5.3cm]{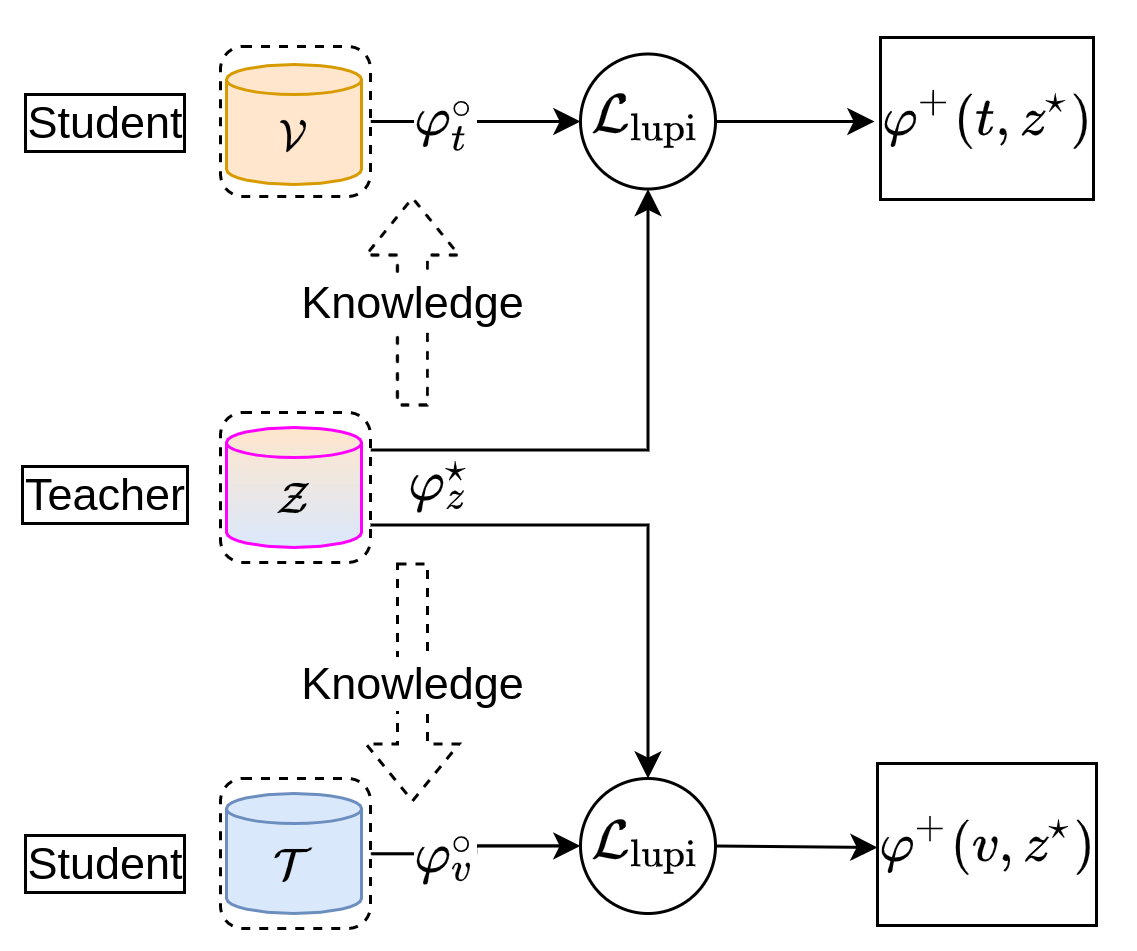}}}  \\
(a) &                    \\ 
{\includegraphics[width=5.3cm]{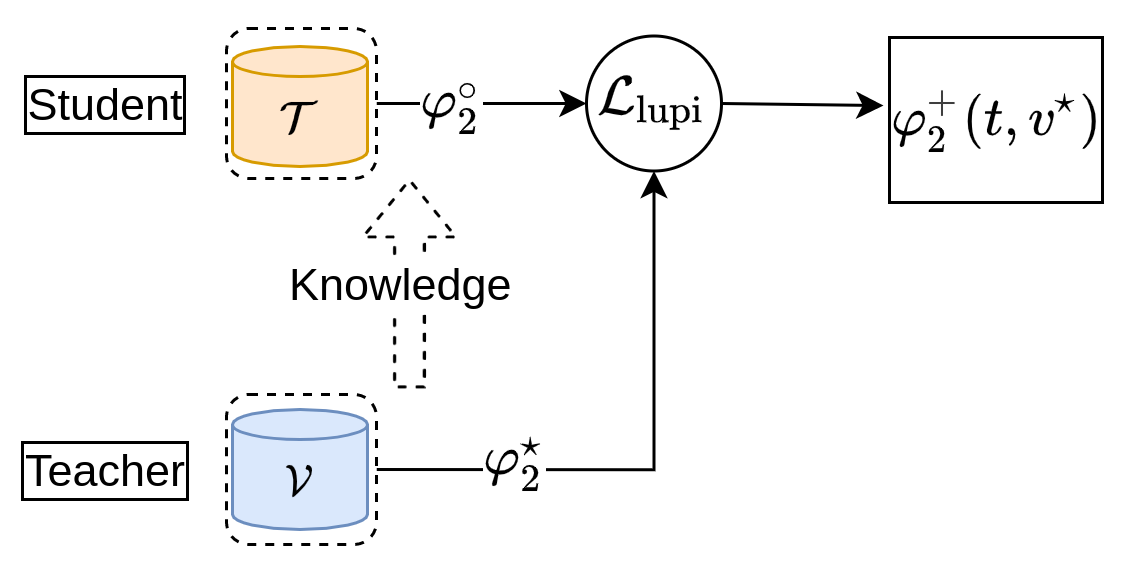}} &                    \\ 
(b)  &  (c)                 \\ 
\end{tabular}
\caption{Summary of methodology to train a model under privileged information: (a) learning under privileged infrared domain, (b) learning under privileged visual domain, and (c) learning under privileged intermediate domain, between infrared and visual.}
\label{fig:our_lupi_approach}
\end{figure}

Two methods were provided to incorporate Privilege Information (PI) through the training process. The first approach uses the infrared modality as PI to train the visible models, as shown in Fig. \ref{fig:our_lupi_approach}.a and the first part of Equ. \ref{eq:visual_lupi}, or vice versa, as shown in Fig. \ref{fig:our_lupi_approach}.b and the second part of Equ. \ref{eq:visual_lupi}.
\begin{equation} \label{eq:visual_lupi}
\begin{aligned}
    \min_{\Theta_1} \mathbf{E}_{v,t^\star,y \sim p(v,t^\star,y)}[\mathcal{L}(\varphi_1^ +(v,t^\star;\Theta_1), y)],
    \\
    \min_{\Theta_2} \mathbf{E}_{t,v^\star,y \sim p(t,v^\star,y)}[\mathcal{L}(\varphi_2^+ (t,v^\star;\Theta_2), y)].
\end{aligned}
\end{equation}
where $\mathcal{L}$ is the loss function. For the second approach, since the domain discrepancies between infrared and visible are numerous \cite{HAT}, we employ a mutual (intermediate) modality while learning a joint model for visible-infrared sensor data (Fig. \ref{fig:our_lupi_approach}.c and the Equ. \ref{eq:z_lupi}). The intermediate modality allows bridging the cross-modality gap by making each main modality benefit from the other more easily, since mutual domain has more information in common with both main domains than each main domains.
\begin{equation} \label{eq:z_lupi}
\begin{aligned}
    \min_{\Theta} \mathbf{E}_{v, t, z^\star,y \sim p(v, t, z^\star,y)}[\mathcal{L}(\varphi^+ (v, t, z^\star;\Theta), y)].
\end{aligned}
\end{equation}
where $z^\star$ is intermediate domain, $\Theta$ is parameter set of the model. 

For these approaches, the teacher's knowledge (related to PI knowledge) is transferred to the student(s) thanks to shared parameter layers after the first convolutional blocks. For instance, the intermediate modality backbone shares all its parameters with the Visible and Infrared modality streams except for the first convolutional block. In addition, specific losses are used to help the knowledge transfer from our specific intermediate modality.

    

%

\subsection{Intermediate Domain Generation}

\begin{figure}[!t] \label{eq:intermediate_generation}
\centering
\begin{tabular}{cc}
 {\includegraphics[width=3cm]{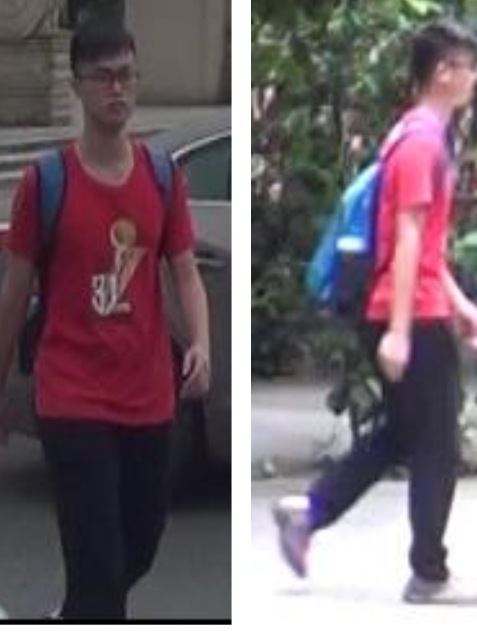}}   & {\includegraphics[width=3cm]{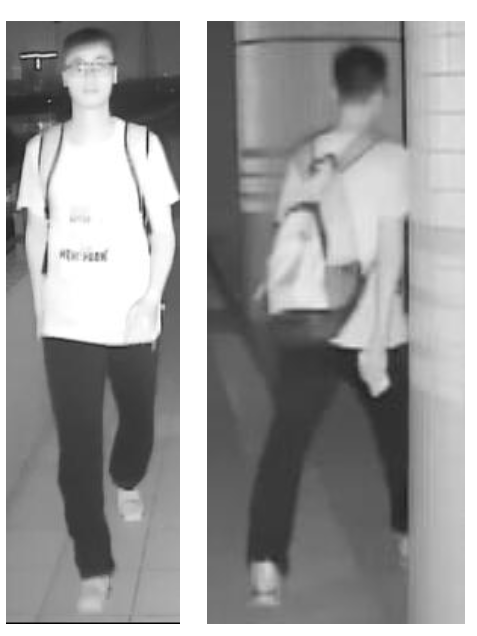}}                      \\
(a) & (b)                  \\
\multicolumn{2}{c}{{\includegraphics[width=6cm]{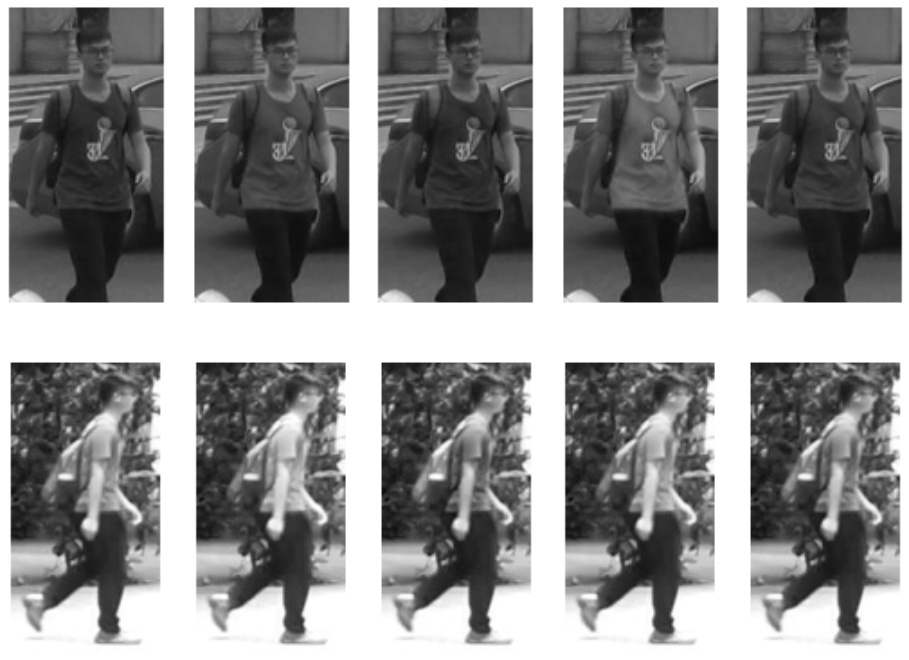}}}       \\
\multicolumn{2}{c}{(c)}   
\end{tabular}
\caption{Illustration domain shift between visible and infrared images: (a) visual images, (b) infrared images, and (c) a selected random linear combination of RGB channels.}
\label{fig:CE-TR}
\end{figure}

Employing an intermediate domain from which the data from the main domains (RGB and IR) have a lower domain shift, can help the model learn DL embeddings with shared discriminative features. The most appropriate choice for an intermediate domain in the context of infrared-visible person ReID appears to be grayscale images, since they can be derived directly from the visible images, but also because it intuitively considers the nature of each modality. Infrared cameras have one sensor which measures the wavelength of 1 mm to the nominal red edge of the visible spectrum, while visible images have three channels showing the wavelength of the red, green, and blue spectrum on persons. So, in a classic cross-modal person-ReID model, the DL model must to mine shared abstraction between these pieces of information to re-identify the same person from three visible channels to one infrared channel, or vice-versa. As a motivating example, Fig. \ref{fig:CE-TR}.a and \ref{fig:CE-TR}.b show Visible and Infrared images of the same person, where a large shift can be visually observed. By generating one channel images from a random selection through the visible channels (see Fig. \ref{fig:CE-TR}.c), the domain shift appears to be reduced significantly.

The semantic features extracted by the model from the synthetic images, which contain one modality, such as infrared, but have pose information from the visible images, may be one solution for modality-independent features between visible and infrared since they rely less on color. Based on such information, synthetic images in the training phase can be seen as the PI, $z^\star$ in Equ. \ref{eq:z_lupi}. 
We trained the model shown in Fig. \ref{fig:our_lupi_approach}.c with the one-channel images used as teacher, and showed that the random selection of images from the color channel (see Equ. \ref{eq:channel_random_selection}) can be an relevant intermediate domain between infrared and visible:
\begin{equation}\label{eq:channel_random_selection}
G_s = \frac{\alpha R+\beta G+ \gamma B}{\alpha+\beta+\gamma} \\
\end{equation}
where $\alpha, \beta, \mbox{and} \gamma \sim  \mathbb{U}(0,1)$, a uniform random distribution generator, and $R$, $G$, and $B$ are one channel color images. Instead of being limited to a specific version of colored images as in \cite{liu2021sfanet,HAT}, our model uses color-free images, without any color information from visible modality, to bridge the gap between visible and infrared image modalities.

\subsection{Loss Functions}

\noindent \textbf{(1) Intermediate Dual Triplet Loss.}
The ultimate goal of deep cross-modal person ReID is to encode an embedding where the same person captured with different camera modalities are more similar in the feature space. In such space, the feature vectors of different individuals should also be as far as possible from each other. Triplet loss is one objective constraint usually used for the model to achieve this goal. This loss compares a reference input (called the anchor) with a matching input person (called positive) and a different input person (called negative). With the intermediate domain, we apply such loss in dual mode for comparing each main domain with the intermediate:  
\begin{equation} \label{eq:gray-triplet}
\begin{split}
    \mathcal{L}_{\text{tri}} =
  \mathbb{E}_{(\mathbf{f}_a,\mathbf{f}_p,\mathbf{f}_n)\in (\mathcal{F}^v, \mathcal{F}^t, \mathcal{F}^z) }[M - D_{\mathbf{f}_a,\mathbf{f}_p} + D_{\mathbf{f}_a,\mathbf{f}_n}] \\
+ \mathbb{E}_{(\mathbf{f}_a,\mathbf{f}_p,\mathbf{f}_n)\in (\mathcal{F}^t, \mathcal{F}^z, \mathcal{F}^v) }[M - D_{\mathbf{f}_a,\mathbf{f}_p} + D_{\mathbf{f}_a,\mathbf{f}_n}] \\
+ \mathbb{E}_{(\mathbf{f}_a,\mathbf{f}_p,\mathbf{f}_n)\in (\mathcal{F}^z, \mathcal{F}^v, \mathcal{F}^t) }[M - D_{\mathbf{f}_a,\mathbf{f}_p} + D_{\mathbf{f}_a,\mathbf{f}_n}] 
\end{split}
\end{equation}
where subscript $a$, $p$, and $n$ stands respectively for anchor, positive and negative indicator. The superscript $t$, $v$, and $z$ select visible, infrared and color-free modalities features.

\noindent \textbf{(2) Color-Free Loss.}
Feature invariance during extraction of the original visible images features and those of the generated modality,  has been shown to be beneficial \cite{HAT}. In this regard, we introduce a color-invariant loss function between visible images and virtual color-free images to enhance the robustness of feature representations  against variations in modality. This color-free modality teaches the visible branch not rely on the colors spectrum. The color-free loss to train the CNN backbone is defined by:
\begin{equation}
\mathcal{L}_c =\mathbb{E}_{(\mathbf{f}_i^v,\mathbf{f}_i^z)\in (\mathcal{F}^v, \mathcal{F}^z) } [\begin{cases}
                    \lvert \mathbf{f}^v_i - \mathbf{f}^z_i \rvert, & \text{if $\lvert \mathbf{f}^v_i - \mathbf{f}^z_i \rvert$} > 0.5\\
                    \alpha_c KL(\mathcal{C}(\mathbf{f}^v_i),\mathcal{C}(\mathbf{f}^z_i)), & \text{otherwise}.
                \end{cases}]
\end{equation}
where $\mathbf{f}^z_i$ is the feature vector extracted from generated images $z_i$, $\mathcal{C}$ is the fully-connected layer that produces the classification probability for each identify of image $i$, and $KL(\cdot)$ is the Kullback-Leibler divergence between two probabilities. Since the augmentation mechanism uses different distortions for RGB and intermediate images, the model may not extract similar features from these types of images, so we use the distance of identification probability between them.

\noindent \textbf{(3) Overall Loss for LUPI.}
The loss proposed to leverage a privileged intermediate domain $\mathcal{L}_\text{lupi}$ is defined by combining $\mathcal{L}_c$ and $\mathcal{L}_{\text{tri}}$:
\begin{equation}
\mathcal{L}_\text{lupi} = \mathcal{L}_{\text{tri}} + \lambda \mathcal{L}_c,
\end{equation}
where $\lambda$ is a predefined trade off parameter to balance the color-free loss. Also, to extract features that are discriminant according to person identity, the total loss $\mathcal{L}$ in Equ. \ref{eq:z_lupi} includes an identity loss:
\begin{equation}
\mathcal{L} = \mathcal{L}_{\text{lupi}} + \mathcal{L}_{\text{id}}. 
\end{equation}
By optimizing DL model parameters with identity supervision, using an $\mathcal{L}_{\text{id}}$  like cross-entropy \cite{all-survey} allows for feature representations that are  identity-variant.

\section{Results and Discussion}

\subsection{Experimental Methodology}

\noindent \textbf{Datasets:}
The two datasets RegDB \cite{regDB}, and SYSU-MM01\cite{SYSU} have been widely used in cross-modal research. SYSU-MM01 is a large-scale dataset containing 22,258 visible images and 11,909 infrared images (captured from four RBG and two near-infrared cameras) of 491 individuals. Following its authors, 395 identities were used as the training set, and 96 identities were used as the testing set for fairness comparisons with other papers. It contains two evaluation modes in SYSU-MM01 based on images in the gallery: single-shot and multi-shot. The former setting allows us to select a random image from each person's identity within the gallery, while the latter setting allows us to select ten images from each identity. The RegDB dataset consists of 4,120 visible and 4,120 infrared images representing 412 identities, which are collected from a single visible and an infrared camera co-located. Each identity contains ten visible images and ten infrared images from consecutive clips. Additionally, ten trail configurations randomly divide the images into two identical sets (206 identifiers) for training and testing. Two different test settings are used -- matching infrared (query) with visible (gallery), and vice versa.

\noindent \textbf{Performance Measures:}
We use the Cumulative Matching Characteristics (CMC), and Mean Average Precision (mAP) as evaluation metrics.

\noindent \textbf{Implementation Details:}
Our proposed approach is implemented using the PyTorch framework. For fair comparison with the other methods, the ResNet50 \cite{resnet} CNN was used as backbone for feature extraction, and pre-trained on ImageNet \cite{imagenet}. Our training process and model was implemented on the code released by \cite{all-survey}, following the same hyper-parameter configuration. For Eq. \ref{eq:intermediate_generation}, we generate three uniform random numbers, and then normalize them for each channel of visible images to create color-free images. At first, all the input images are resized to 288 × 144, then cropped randomly, and filled with zero-padding or with the mean of pixels as a data augmentation approach. A stochastic gradient descent (SGD) optimizer is used in the optimization process. 
The used margin used for the triplet loss is the same as \cite{all-survey}, and the warming-up strategy is applied in the first ten epochs. In order to perform the triplet loss in each mini-batch during training, the group must contain at least two distinct individuals for whom at least two images (one for anchor and one for positive) have to be included. So, in each training batch, we select $b_s=8$ persons from the dataset and $n_p=4$ positive images for each person, still following \cite{all-survey} parameters. The $\alpha_c=0.5$ and $\lambda=10$ is set.

\subsection{Comparison with the State-of-Art}
Tables \ref{tab:all-results} and \ref{tab:all-results-RegDB} compare the results with our proposed approach against state-of-the-art cross-modal VI-ReID methods published in the past two years, on the SYSU-MM01 and RegDB datasets. The results show that our method outperforms these methods in various settings. Our method does not require additional and costly image generation, conversion from visible to infrared, or adversarial learning process. It is very to generate random channel images using matrix dot production, which is a simple mathematics operation with low computational overhead. Also, our model learns a robust feature  representation against different cross-modality matching settings. Notably, on the large-scale SYSU-MM01 dataset, our model improves the Rank-1 accuracy and the mAP score by 23.58\% and 19.11\%, respectively, over the AGW model \cite{all-survey} (our baseline). Recall that our approach can be used on top of any cross-modal approach, so the performances obtained working over the baseline are promising. On SYSU-MM01, our model shows an improvement of 5.61\% in terms of Rank-1 and of 1.78\% mAP in comparison to the second best approach considering Indoor Search scenario. For RegDB dataset, from Visible to Thermal ReID, our model improves the best R1 and mAP by respectively 2.92\% and 1.80 \%. Similar improvement can be observed from the RegDB Thermal to Visible setting. Another advantage of our approach is that it can be combined with other state-of-art models without any overhead during inference, since it only triggers training process to force visible features such that they are less discriminative on colors. 

\begin{table}[!t]
\centering
\scalebox{0.9}{%
\begin{tabular}{|c|l||c|c|c||c|c|c|} 
\hline
\multicolumn{2}{|c||}{\textbf{Family}}  & \multicolumn{3}{c||}{All Search}  & \multicolumn{3}{c|}{Indoor Search} \\ \hline
\multicolumn{2}{|c||}{\textbf{Method}}   & \textbf{R1(\%)} & \textbf{R10(\%)} & \textbf{mAP(\%)} & \textbf{R1(\%)} & \textbf{R10(\%)} & \textbf{mAP(\%)} \\ \hline \hline
\multirow{15}{*}{\rotatebox[origin=c]{90}{Representation}} &  DZP\cite{DZP}  & 14.8  & 54.1   & 16.0 & 20.58 & 68.38 & 26.92 \\ 
\cline{2-8}
  &  CDP\cite{Fan2020CrossSpectrumDP}   & 38.0~ & 82.3   & 38.4 & - & - & -\\ 
\cline{2-8}
  &  SFANET\cite{liu2021sfanet}    & 65.74  & 92.98  & 60.83 & 71.60 & 96.60 & 80.05 \\ 
\cline{2-8}
  &  HAT\cite{HAT}  & 55.29 & 92.14  & 53.89 & - & - & -\\ 
\cline{2-8}
  &  xModal \cite{li2020infrared}    & 49.92 & 89.79  & 50.73 & 62.1 & 95.75 & 69.37 \\ 
\cline{2-8}
&  CAJ \cite{ye2021channel}   & 69.88 & -  & 66.89  & 76.26 & 97.88 & 80.37\\ 
\cline{2-8}

&  EDFL \cite{EDFL:journals/corr/abs-1907-09659}      & 36.94     & 84.52  & 40.77 & - & - & -      \\ 
\cline{2-8}
& BDTR \cite{Bi-Di_Center-Constrained}     & 27.82 & 67.34  & 28.42 & 32.46 & 77.42 & 42.46     \\ 
\cline{2-8}
& AGW\cite{all-survey}  & 47.50 & - & 47.65  & 54.17 & 91.14 & 62.97\\
\cline{2-8}
& cmGAN \cite{cm-gan}    & 26.97   & 67.51 & 31.49 & 31.63 & 77.23 & 42.19 \\ 
\cline{2-8}
& MANN \cite{adv-modal}   & 30.04 & 74.34  & 32.18 & - & - & - \\
\cline{2-8}

& TSLFN  \cite{hetero-center}   & 62.09  & 93.74  & 48.02 & 59.74 & 92.07 & 64.91  \\ 
\cline{2-8}
& DDAG \cite{DDAG}     & 54.75 & 90.39  & 53.02 & 61.02 & 94.06 & 67.98      \\
\cline{2-8}
& JFLN \cite{JFLN}     & 66.11 & 95.69  & 64.93 & - & - & -   \\
\cline{2-8}
& MPANet \cite{wu2021Nuances}  & 70.58 & 96.21  & \textbf{68.24} & 76.74 & 98.21 & 80.95 \\

\hline
\hline
\multirow{5}{*}{\rotatebox[origin=c]{90}{Translation}}   & AGAN  \cite{Wang_2019_ICCV_AlignGAN}  & 42.4   & 85.0 & 40.7 & 45.9 & 92.7 & 45.3 \\ 
\cline{2-8}
& D\textsuperscript{2}RL  \cite{D2RL}  & 28.9 & 70.6 &  29.2 & - & - & - \\ 
\cline{2-8}
 & JSIA \cite{JSIA-paired-images1}   & 45.01  &  85.7 & 29.5  & 43.8 & 86.2 & 52.9\\ 
\cline{2-8}
 & Hi-CMD \cite{HI-CMD}  & 34.94   & 77.58 & 35.94 & - & - & -  \\
\cline{2-8}
 & TS-GAN \cite{zhang2021rgb}   & 58.3   & 87.8   & 55.1 & 62.1 & 90.8 & 71.3  \\ 
\hline
\hline
\multirow{6}{*}{\rotatebox[origin=c]{90}{Co-Learning}}   & HCML \cite{HCML}  & 14.32 & 53.16 & 16.16  & - & - & - \\ 
\cline{2-8}
& DHML \cite{zhang2019dhml}  &  60.08 & 90.89 & 47.65 & 56.30 & 91.46 & 63.70  \\ 
\cline{2-8}
& SSFT \cite{cmSSFT}  &  63.4 & 91.2 & 62.0  & 70.50 & 94.90 & 72.60   \\ 
\cline{2-8}
& SIM \cite{SIMcm} &  56.93 & - & 60.88 & - & - & - \\ 
\cline{2-8}
& LbA\cite{park2021learning}  &  55.41  &   & 54.14  & 61.02 & - & 66.33  \\ 
\cline{2-8}
& CM-NAS\cite{fu2021cm}  &  61.99  & 92.87  &  60.02 & 67.01 & 97.02 & 72.95 \\
\hline
\hline
\multicolumn{2}{|c||}{Ours} & \textbf{71.08} & \textbf{96.42} & 67.56 & \textbf{82.35} & \textbf{98.3} & \textbf{82.73}\\
\hline
\end{tabular}
}
\caption{Performance of state-of-art techniques for cross-modal person ReID on the SYSU-MM01 dataset.}
\label{tab:all-results}
\end{table}

\begin{table}[!h]
\centering
\scalebox{0.9}{%
\begin{tabular}{|c|l||c|c|c||c|c|c|} 
\hline
\multicolumn{2}{|c||}{\textbf{Family}}  & \multicolumn{3}{c||}{Visible $\rightarrow$ Thermal}  & \multicolumn{3}{c|}{Thermal $\rightarrow$ Visible} \\ \hline
 \multicolumn{2}{|c||}{\textbf{Method}}   & \textbf{R1(\%)} & \textbf{R10(\%)} & \textbf{mAP(\%)} & \textbf{R1(\%)} & \textbf{R10(\%)} & \textbf{mAP(\%)}    \\  \hline \hline
\multirow{12}{*}{\rotatebox[origin=c]{90}{Representation}} &  DZP\cite{DZP}  & 17.75  & 34.21   & 18.90 & 16.63 & 34.68 & 17.82 \\ 
\cline{2-8}
  &  CDP\cite{Fan2020CrossSpectrumDP}   & 65.3 & 84.5   & 62.1 & 64.40 & 84.50 & 61.50\\ 
\cline{2-8}
  &  SFANET\cite{liu2021sfanet}    & 76.31  & 91.02  & 68.00 & 70.15 & 85.24 & 63.77 \\ 
\cline{2-8}
  &  HAT\cite{HAT}  & 71.83 & 87.16  & 67.56 & 70.02 & 86.45 & 66.30\\ 
\cline{2-8}
  &  xModal \cite{li2020infrared}    & 62.21 & 83.13  & 60.18 & - & - & - \\ 
\cline{2-8}
&  CAJ \cite{ye2021channel}   & 85.03 & 95.49  & 79.14  & 84.75 & 95.33 & 77.82\\ 
\cline{2-8}

&  EDFL \cite{EDFL:journals/corr/abs-1907-09659} & 48.43 & 70.32 & 48.67 & 51.89 & 72.09  & 52.13      \\ 
\cline{2-8}
& BDTR \cite{Bi-Di_Center-Constrained}     & 33.47 & 58.96  & 31.83 & 34.21 & 58.74 & 32.49     \\ 
\cline{2-8}
& AGW\cite{all-survey}  & 70.05 & - & 50.19  & 70.49 & 87.21 & 65.90\\
\cline{2-8}
& MANN \cite{adv-modal}   & 48.67 & 71.55 & 41.11 & 38.68 & 60.82  & 32.61\\
\cline{2-8}

& DDAG \cite{DDAG}     & 69.34 & 86.19& 63.46 & 68.86 & 85.15  & 61.80      \\
\cline{2-8}
& MPANet \cite{wu2021Nuances}  & 83.70  & - & 80.9 & 82.8 & - & 80.7 \\ 
\hline
\hline
\multirow{5}{*}{\rotatebox[origin=c]{90}{Translation}}   & AGAN  \cite{Wang_2019_ICCV_AlignGAN}  & 57.9   & - & 53.6 & 56.3 & - & 53.4 \\ 
\cline{2-8}
& D\textsuperscript{2}RL  \cite{D2RL}  & 43.4 & 66.1 &  44.1 & - & - & - \\ 
\cline{2-8}
 & JSIA \cite{JSIA-paired-images1}   & 48.5  &  - & 49.3  & 48.1 & - & 48.9\\ 
\cline{2-8}
 & Hi-CMD \cite{HI-CMD}  & 70.93   & 86.39 & 66.04 & - & - & -  \\
\cline{2-8}
 & TS-GAN \cite{zhang2021rgb}   & 68.2   & -   & 69.4 & - & - & -  \\ 
\hline
\hline
\multirow{5}{*}{\rotatebox[origin=c]{90}{Co-Learning}}   & HCML \cite{HCML}  & 24.44  &  47.53 & 20.80  & 21.70 & 45.02 & 22.24 \\ 
\cline{2-8}
& SSFT \cite{cmSSFT}  &  71.0 & - & 71.7  & - & - & -   \\ 
\cline{2-8}
& SIM \cite{SIMcm} &  74.47 & - & 75.29 & 75.24 & - & 78.30 \\ 
\cline{2-8}
& LbA\cite{park2021learning}  &  74.17 & - & 67.64 & 72.43  & -  & 65.64  \\ 
\cline{2-8}
& CM-NAS\cite{fu2021cm}  &   84.54 & 95.18 & 80.32 & 82.57  & 94.51  &  78.31 \\
\hline
\hline
\multicolumn{2}{|c||}{Ours} & \textbf{87.95} & \textbf{98.3} & \textbf{82.73} & \textbf{86.80} & \textbf{96.02} & \textbf{81.26}\\
\hline
\end{tabular}
}
\caption{Performance of state-of-art techniques for cross-modal person ReID on the RegDB dataset.}
\label{tab:all-results-RegDB}
\end{table}

\subsection{Ablation Study} \label{sec:ablation_study}
In this section, impact on performance of each component in our model is analyzed. At first, it begins with domain shift between visible and infrared images, then comparing the loss functions on training process. Also the qualitative results are discussed in the supplementary. 

\noindent \textbf{Domain Shift between Visual, Infrared and Gray.}
In a way to determine the domain shift between domains, we explored two distinct strategies, the first Maximum Mean Discrepancy (MMD) \cite{gretton2012kernelMMD} which is done in supplementary and then measuring and comparing mAP and Rank-1 directly.
We compared such metrics to measure the domain shift on model performance, by playing with different modalities as query and gallery in the inference phase while training only from each modality (three first columns in Table \ref{tab:gray-modal}). To do so, we used the AGW model \cite{all-survey} with its default settings since its architecture corresponds to our final model, except we used only a single branch here by which we fed the query and gallery images. Indeed, AGW used two Resnet50 with shared parameters between the visible and infrared branches for the cross-modal ReID purpose, which should not be used this way here since, in the inference phase, we need to feed the cross-modal images from one branch. 

In practice, we observe that the model trained with only gray images performs better on other modalities. For instance, the mAP of the \textbf{G}ray model is 5\% better than the \textbf{V}isible only model when working from infrared images (Respectively 27.05\% and 21.97\%). Similarly, the \textbf{G}ray model is ahead of the \textbf{I}nfrared one by a large margin when performing from visible images, respectively 78.60\% and 20.24\% mAP. Those results confirm the intermediate position of the gray domain in between the two main domains. Moreover, this model has better performance to retrieve the best candidate (R1 Accuracy) when the Query comes from \textbf{I}nfrared and Gallery from \textbf{V}isual in comparison to model trained only with \textbf{V}isuals (11.2 to 9.86). 

\begin{table}[!t]
\centering
\scalebox{0.95}{
\begin{tabular}{|c||c|c||c|c||c|c||c|c||c|c|}
\hline
\multirow{2}{*}{\textbf{Testing}} & \multicolumn{10}{c|}{\textbf{Training Modalities}}  \\ \cline{2-11}
                                 & \multicolumn{2}{c||}{V}         & \multicolumn{2}{c||}{I}      & \multicolumn{2}{c||}{G}         & \multicolumn{2}{c||}{I-V}      & \multicolumn{2}{c|}{I-V-G}       \\ \cline{2-11}
Query $\rightarrow$ Gallery        & \textbf{R1}    & \textbf{mAP} & \textbf{R1} & \textbf{mAP} & \textbf{R1}    & \textbf{mAP} & \textbf{R1} & \textbf{mAP}   & \textbf{R1}    & \textbf{mAP}   \\ \hline \hline
V $\rightarrow$ V          & \textbf{97.95} & 91.67        & 60.61       & 20.24        & 95.69          & 78.60        & 97.40       & \textbf{91.82} & 97.22          & 89.22          \\ \hline
I$ \rightarrow$ I          & 49.76          & 21.97        & 94.90       & 80.38        & 56.93          & 27.05        & 95.96       & 80.49          & \textbf{96.92} & \textbf{82.78} \\ \hline
G $\rightarrow$ G          & 88.36          & 54.26        & 49.30       & 16.30        & \textbf{96.10} & 80.06        & 90.16       & 60.35          & 95.94          & \textbf{84.16} \\ \hline
I$ \rightarrow$ V          & 9.86           & 6.83         & 15.72       & 5.92         & 14.20          & 9.01         & 58.69       & 41.57          & \textbf{65.16} & \textbf{46.94} \\ \hline
I $\rightarrow$ G          & 10.89          & 6.50         & 13.09       & 5.71         & 16.15          & 9.22         & 50.85       & 27.79          & \textbf{64.98} & \textbf{45.01} \\ \hline
\end{tabular}

}
\caption{Results of uni-modal and multi-modal training on Visual (V), Infrared (I), and Grayscale (G) modality images of the SYSU-MM01 dataset under the multi-shot setting.  G images are grayscale of V images. During testing, query and gallery images are processed by the model separately. Note that when the query and gallery pairs are in the same modality, the same cameras are excluded to avoid the existence of images from in the same scene.}
\label{tab:gray-modal}
\end{table}

\noindent \textbf{Effect of Random Gray Image Generation.}
As the visual differences between gray and infrared images are much lower than between visible and infrared images in our eyes, we take advantage of such phenomena by learning our model with the gray version of colored images alongside infrared and visible images. In fact, instead of directly optimizing the distances between infrared and visible, the model utilizes gray images as a bridge, and tries to represent infrared and visible features so that they are similar to the gray ones. The result are shown Table \ref{tab:gray-modal}. The main result in the table concerns the ability of the models to well re-identify from \textbf{I}nfrared to \textbf{V}isible. Here, the gray model version (\textbf{I}-\textbf{V}-\textbf{G} training) is successfully ahead of the model trained with infrared and visible images only (\textbf{I}-{V}), reaching 46.94\% against 41.57\% mAP respectively. Also, the gray model is much better to translate from and to the gray domain, which was expected but is of good omen regarding the intermediate characterisation of this domain. Finally, we can also observe pretty equivalent performances under \textbf{V} to \textbf{V} and \textbf{I} to \textbf{I} testing settings, making the gray version the way to go for a robust person ReID. Additionally, in comparison with \textbf{V} training and \textbf{I} training (Table \ref{tab:gray-modal}), which are respectively lower and upper-bound, the performance of our approach is close to the upper-bound, showing that the model can benefit from the privileged missed modality in LUPI paradigm.  
\begin{table}[!h]
\centering
\scalebox{0.9}{
\begin{tabular}{|l||cc|cc|c|}
\hline
\multirow{2}{*}{\textbf{Model}} & \multicolumn{2}{c|}{\textbf{R1(\%)}}                 & \multicolumn{2}{c|}{\textbf{mAP(\%)}}                & \multirow{2}{*}{\textbf{Memory}} \\ \cline{2-5}
                       & \multicolumn{1}{l|}{ Single-Shot }    & Multi-Shot     & \multicolumn{1}{l|}{ Single-Shot }    & Multi-Shot     &                             \\ \hline \hline
Pre-trained            & \multicolumn{1}{c|}{2.26}           & 3.31           & \multicolumn{1}{c|}{3.61}           & 1.71           & 23.5M                       \\ \hline
Visible-trained        & \multicolumn{1}{c|}{7.41}           & 9.86           & \multicolumn{1}{c|}{8.83}           & 6.83           & 23.5M                       \\ \hline
Baseline               & \multicolumn{1}{c|}{49.41}          & 58.69          & \multicolumn{1}{c|}{41.55}          & 38.17          & 23.5M                       \\ \hline
Grayscale                   & \multicolumn{1}{c|}{61.45}          & 65.16          & \multicolumn{1}{c|}{59.46}          & 46.94          & 23.7M                       \\ \hline
RandG                  & \multicolumn{1}{c|}{64.50}          & 69.01          & \multicolumn{1}{c|}{61.05}          & 50.84          & 23.7M                       \\ \hline
RandG + Aug              & \multicolumn{1}{c|}{\textbf{71.08}} & \textbf{77.23} & \multicolumn{1}{c|}{\textbf{66.76}} & \textbf{61.53} & 23.7M                       \\ \hline
Infrared (train-test)   & \multicolumn{1}{c|}{90.34}          & 94.90          & \multicolumn{1}{c|}{87.45}          & 80.38          & 23.5M                       \\ \hline
\end{tabular}
}
\caption{Evaluation of each component on the SYSU-MM01 dataset under single- and multi-shot settings.}
\label{table:cross_SYSU_Result}
\end{table}

\noindent \textbf{Effect of Intermediate Dual Triplet Loss.}
Combined with the dual triplet loss, it has been demonstrated that the bridging cross-modality triplet loss provides strong supervision to improve discrimination in the testing phase. Using such regularization, as shown in Table \ref{table:losses}, improves the mAP by 9 and rank-1 by 12 percent. 

\noindent \textbf{Effect of Color-Free loss.}
A comparison of the results of our model with and without that loss function is conducted in order to verify the effectiveness of the proposed loss function in focusing on color-independent discriminative parts of visible images. Table \ref{table:losses} shows that when the model uses such regularizing loss, it improves the mAP by 8 and rank-1 by 11 percentile points. 

\begin{table}[!t]
\centering
\scalebox{0.95}{
\begin{tabular}{|l||c|c|c|c|c|}
\hline
\textbf{Strategy}        & $\mathcal{L}_c$ & $\mathcal{L}_{tri}$ & \textbf{R1 (\%)} & \textbf{R10(\%)} & \textbf{mAP (\%)} \\ \hline \hline
Baseline         & $\times$        & $\times$          & 42.89            & 83.51            & 43.51             \\ \hline
Triplet Loss    & $\times$        & \checkmark      & 55.56            & 90.40            & 52.96             \\ \hline
Color Loss      & \checkmark    & $\times$          & 53.97            & 89.32            & 51.68             \\ \hline
Ours            & \checkmark    & \checkmark       & 64.50            & 95.19            & 61.05             \\ \hline
\end{tabular}
}
\caption{Impact on performance of loss functions for the SYSU-MM01 dataset.}
\label{table:losses}
\end{table}

\section{Conclusions}
In this paper, we formulated the cross-modal VI-ReID as a problem in learning under privileged information problem, by considering PI as an intermediate domain between infrared and visible modalities. We devised a new method to generate images between visible and infrared domains that provides additional information to train a deep ReID model through intermediate domain adaptation. Furthermore, we show that using randomly generated grayscale images as PI can significantly improve cross-modality recognition. Future directions involve using adaptive image generation based on images from the main modalities, and an end-to-end learning process to find the one channel images as infrared version of visible. Our experiments on the challenging SYSU-MM01 \cite{SYSU} and RegDB \cite{regDB} datasets show that our proposed approach can significantly outperform state-of-art methods specialized for cross-modal VI-ReID. This does not require extra parameters, nor additional computation during inference.

\vspace*{0.2cm}
\noindent \textbf{Acknowledgements:} This work was supported by Nuvoola AI Inc., and the Natural Sciences and Engineering Research Council of Canada.
\newpage

%
\bibliographystyle{splncs04}
\bibliography{\string paper}
\end{document}


\pagestyle{headings}
\mainmatter
\def\ECCVSubNumber{18}  
\def\etal{\emph{et al}.}

\title{Visible-Infrared Person Re-Identification Using Privileged Intermediate Information \\ (Supplementary)} 

\titlerunning{Visible-Infrared Person-ReID Using LUPI}
%
\author{
    Mahdi Alehdaghi\orcidID{0000-0001-6258-8362} \and
    Arthur Josi\orcidID{0000-0002-4362-4898} \and
    Rafael M. O. Cruz\orcidID{0000-0001-9446-1040} \and
    Eric Granger\orcidID{0000-0001-6116-7945}
} 
\authorrunning{M. Alehdaghi et al.}
%
\institute{Laboratoire d’imagerie, de vision et d’intelligence artificielle (LIVIA)\\ 
Dept. of Systems Engineering,  ETS Montreal, Canada\\
\email{\{mahdi.alehdaghi.1, arthur.josi.1\}@ens.etsmtl.ca, \\ \{rafael.menelau-cruz, eric.granger\}@etsmtl.ca}}

\maketitle

\begin{abstract}

\keywords{Person Re-Identification, Cross-Modal Recognition, Multi-Step Domain Adaptation, Learning Under Privileged Information}
\end{abstract}

\section{Results and Discussion}

\subsection{Domain Shift between Visual, Infrared and Gray} \label{sec:domain_shift}
One of ways to determine the domain shift between domains, is using directly global shift (without any knowledge) between them. Also, for showing domain shift between colored and infrared images is larger than color-free images, at first, sine the gray images of visible domain are one version of color-free images, we used gray images, then such color-free ones. For example, we worked with the Maximum Mean Discrepancy (MMD) \cite{gretton2012kernelMMD} metric from the features of the different modalities extracted by a pre-trained ResNet50 on Imagenet in table \ref{tab:MMD_measure}. On both datasets, the domain shift between infrared and gray images is lower than the ones from \textbf{G}ray to the main domains. For example on SYSU-MM01 dataset, \textbf{G}ray and \textbf{I}nfrared are respectively $0.65$ and $0.86$ away of the \textbf{V}isible domain. 
\begin{table}[!h]
\small
\centering
\begin{tabular}{|c||c|c|c||c|c|c|}
\hline
  & \multicolumn{3}{c||}{\textbf{SYSU-MM01}} & \multicolumn{3}{c|}{\textbf{RegDB}} \\ \hline
  Modality & \multicolumn{1}{c|}{I}      & \multicolumn{1}{c|}{V}      & \multicolumn{1}{c||}{G}      & \multicolumn{1}{c|}{I}       & \multicolumn{1}{c|}{V}      & \multicolumn{1}{c|}{G}      \\ \hline  \hline
I & 0      & 0.8610 & 0.6476 & 0       & 1.0551 & 0.9286 \\ \hline
V & 0.8610 & 0      & 0.3429 & 1.0551  & 0      & 0.3902 \\ \hline
\end{tabular}
\caption{Maximum mean discrepancy (MMD) distance between Visual (V), Infrared (I), and Grayscale (G) modality images in SYSU-MM01 and RegDB datasets. G images are the grayscale version of V images.}
\label{tab:MMD_measure}
\end{table}

\subsection{Qualitative Evaluation:}

\begin{figure}[!ht]
\centering
\centerline{
\begin{tblr}{
  colspec = {Q[m]X[c]  X[c]},
}
\setlength\extrarowheight{20pt}

   & Train & Test  \\
(a)
& \raisebox{-.5\height}{\includegraphics[width=5.2cm]{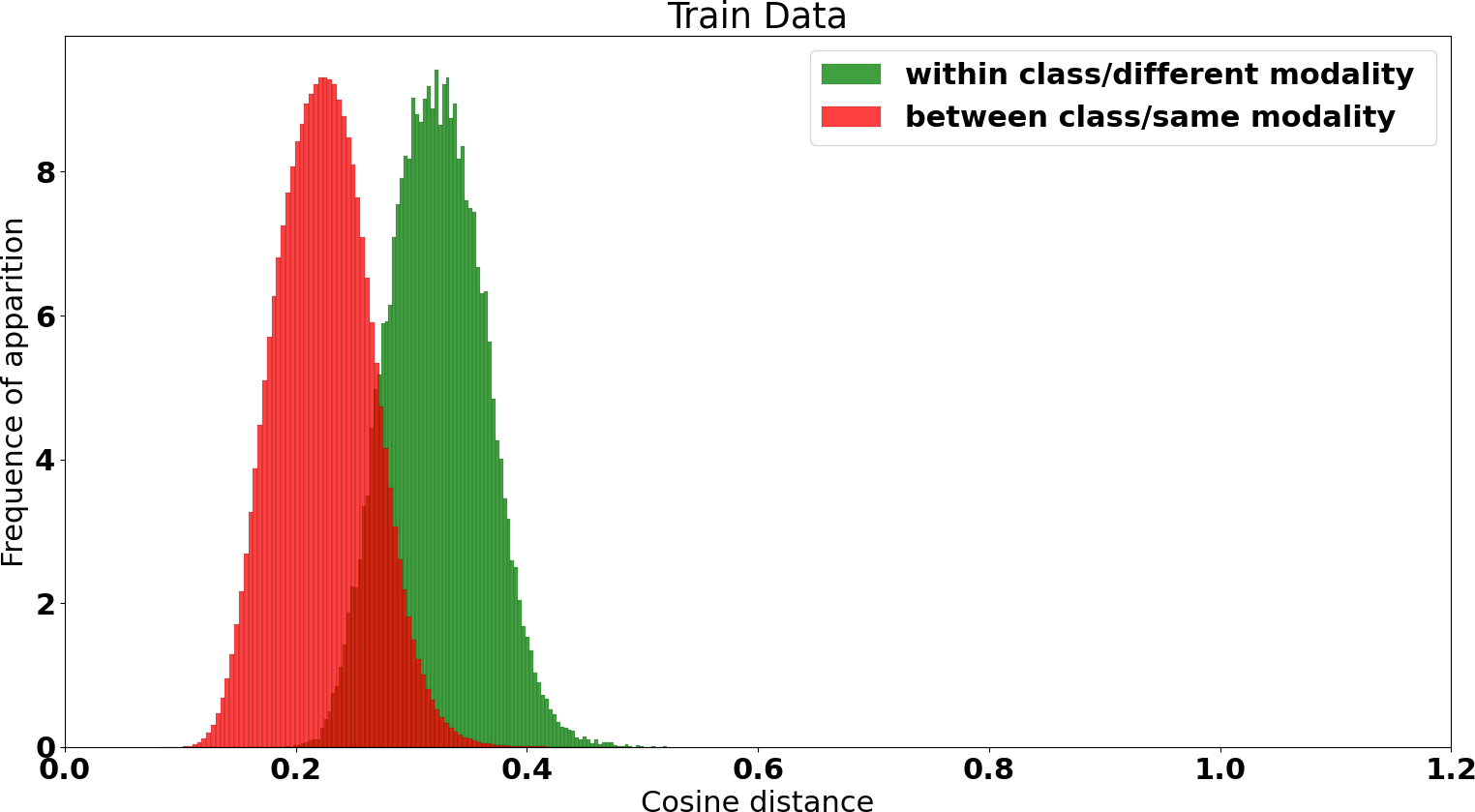}}&\raisebox{-.5\height}{\includegraphics[width=5.2cm]{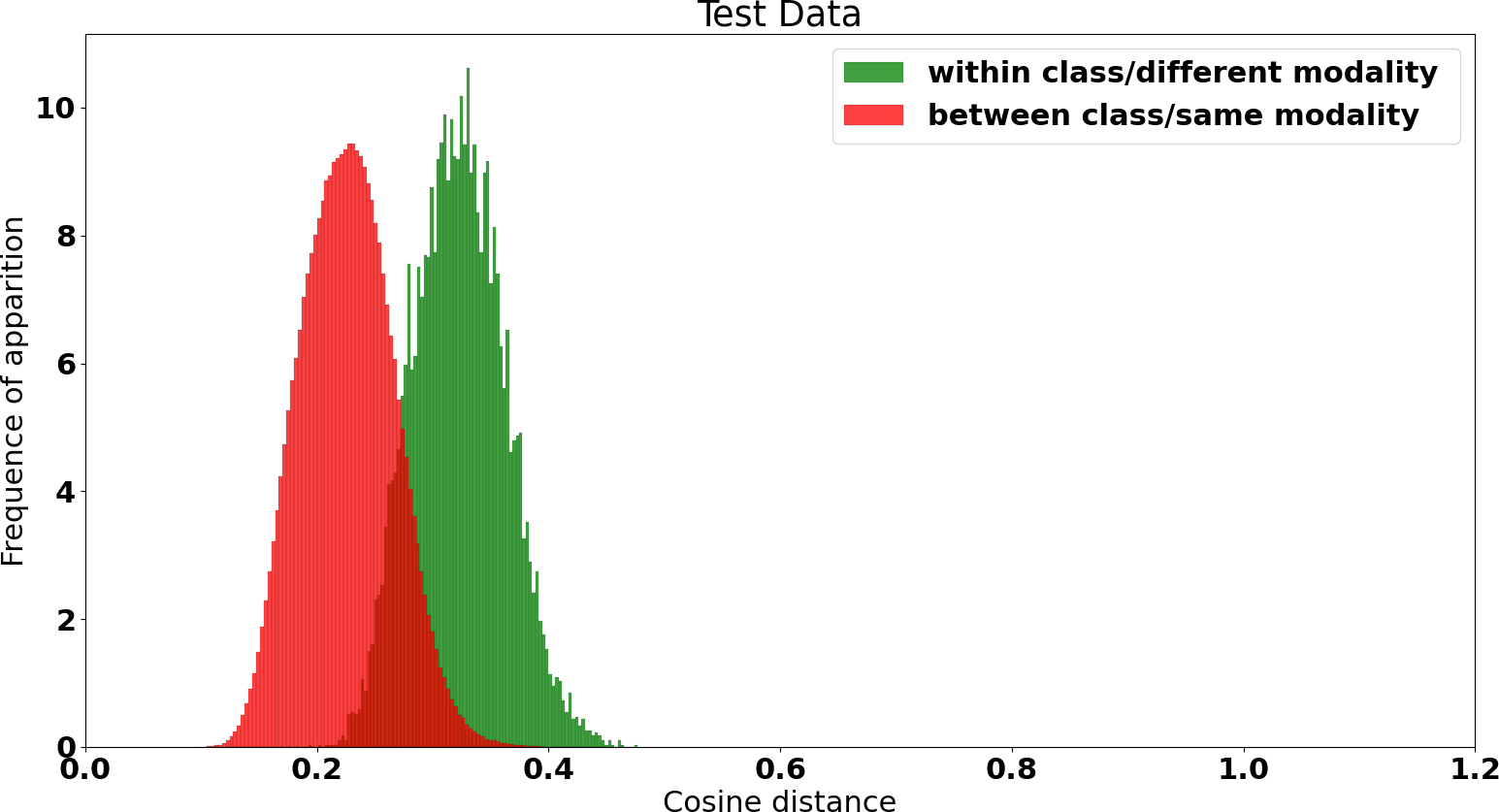}}  \\ 
(b) 
& \raisebox{-.5\height}{\includegraphics[width=5.2cm]{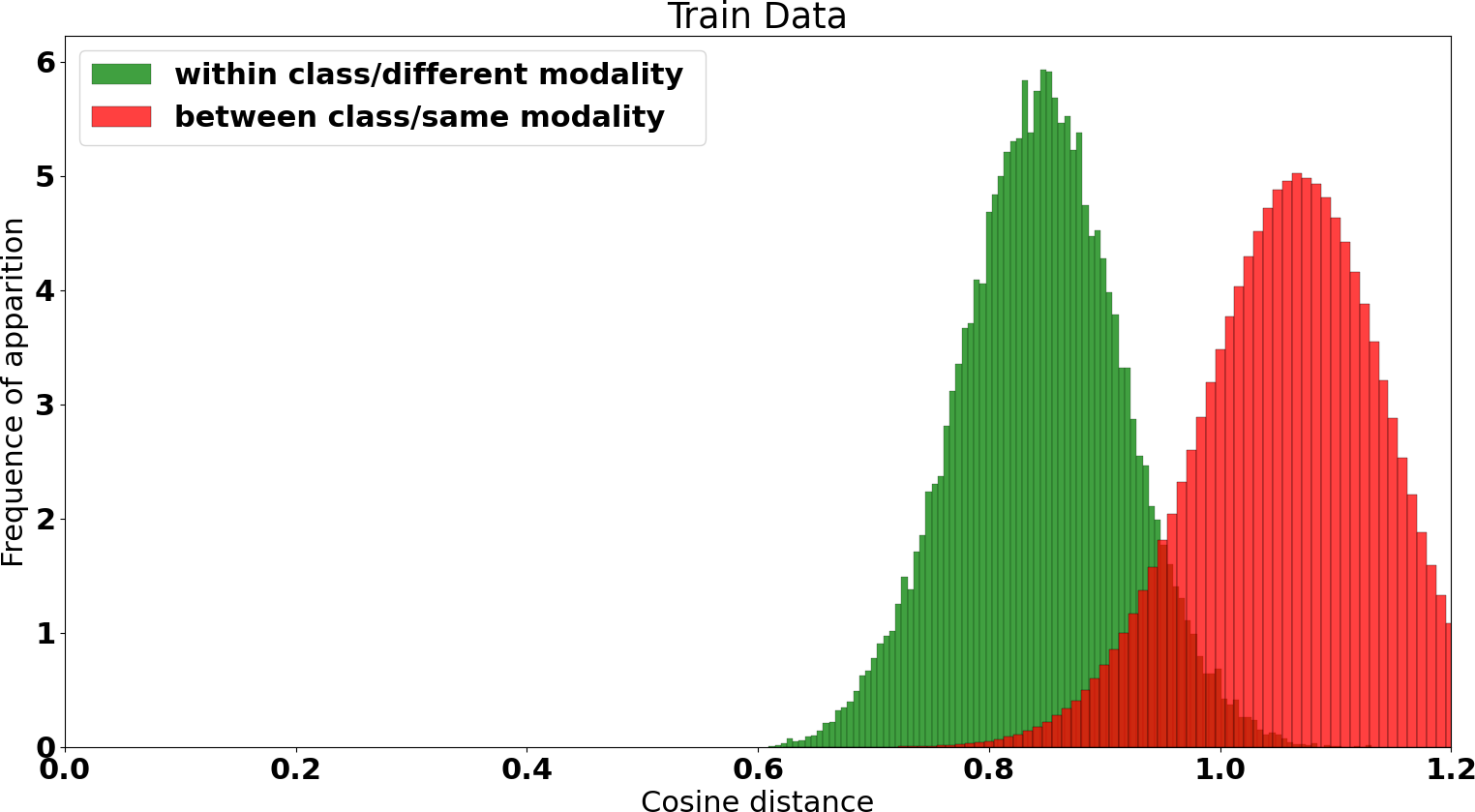}} & \raisebox{-.5\height}{\includegraphics[width=5.2cm]{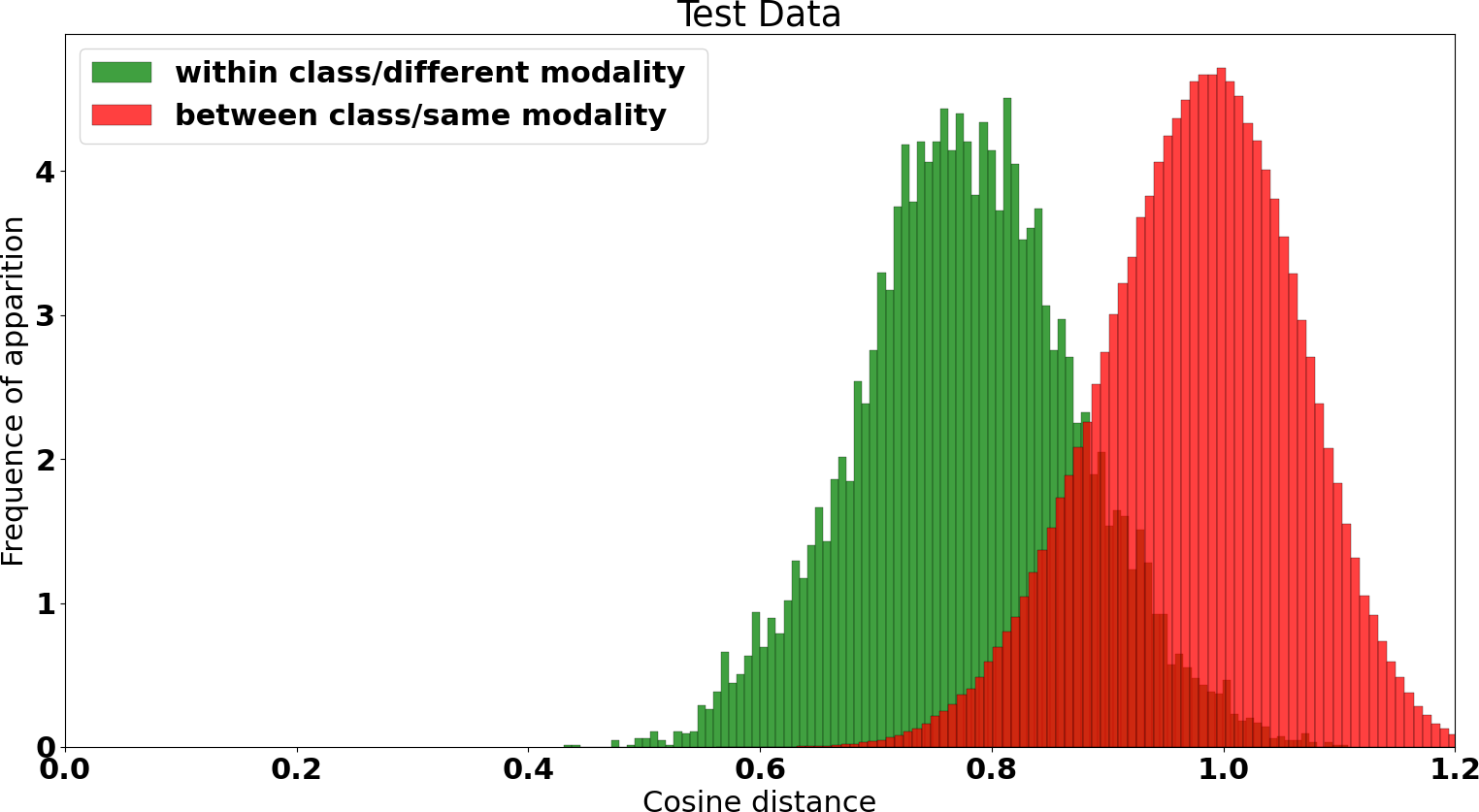}}     \\
(c) 
& \raisebox{-.5\height}{\includegraphics[width=5.2cm]{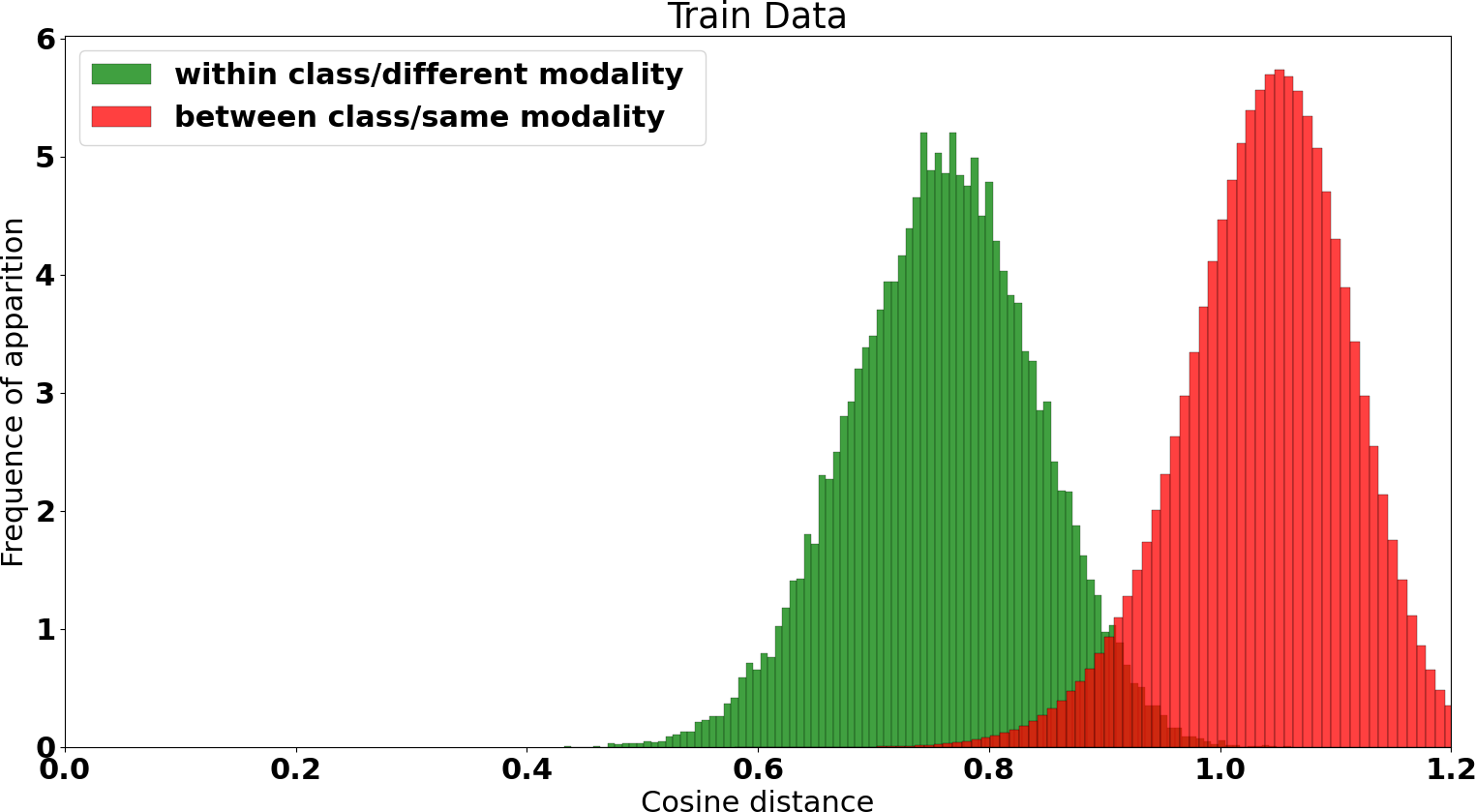}} & \raisebox{-.5\height}{\includegraphics[width=5.2cm]{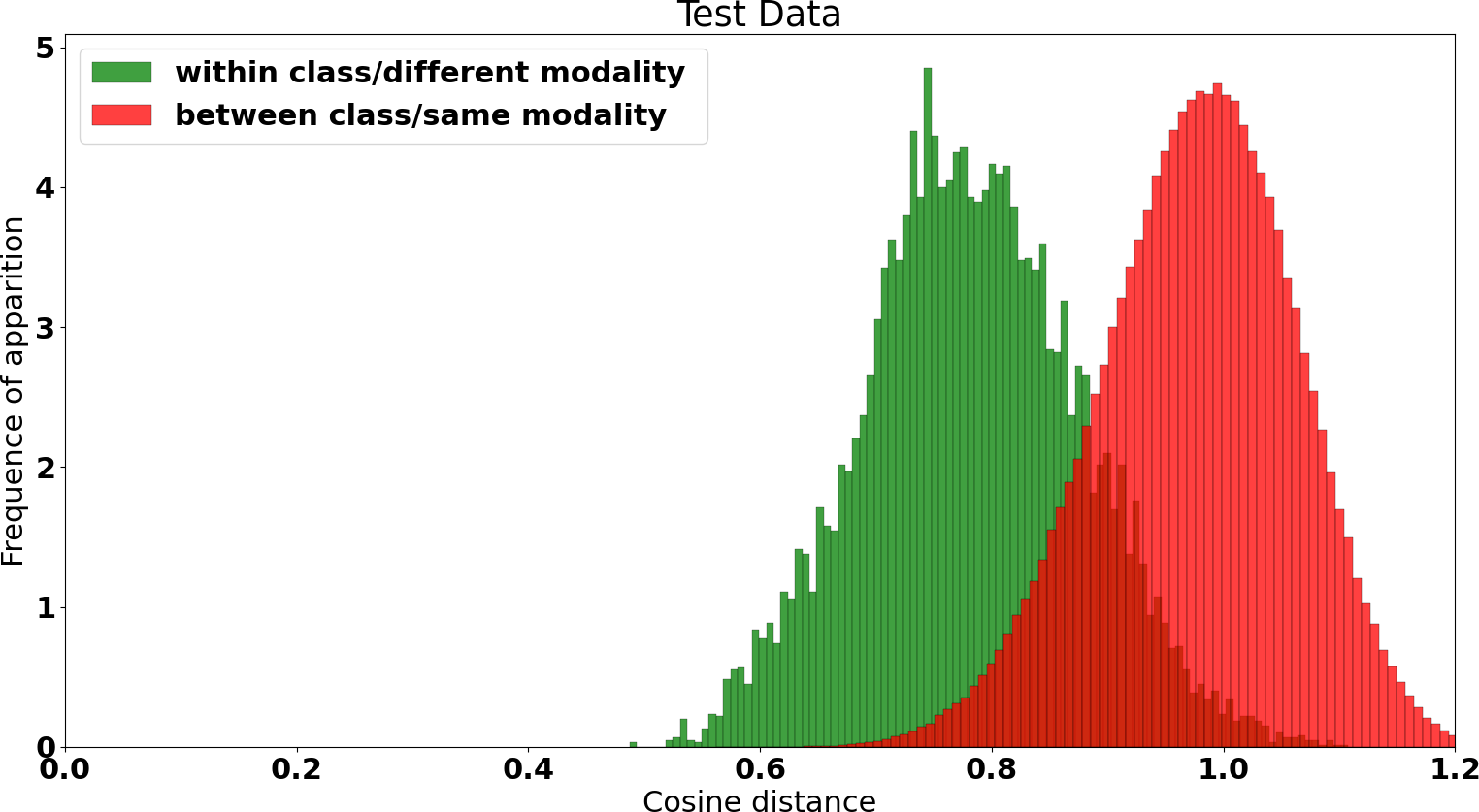}}    \\
(d) 
& \raisebox{-.5\height}{\includegraphics[width=5.2cm]{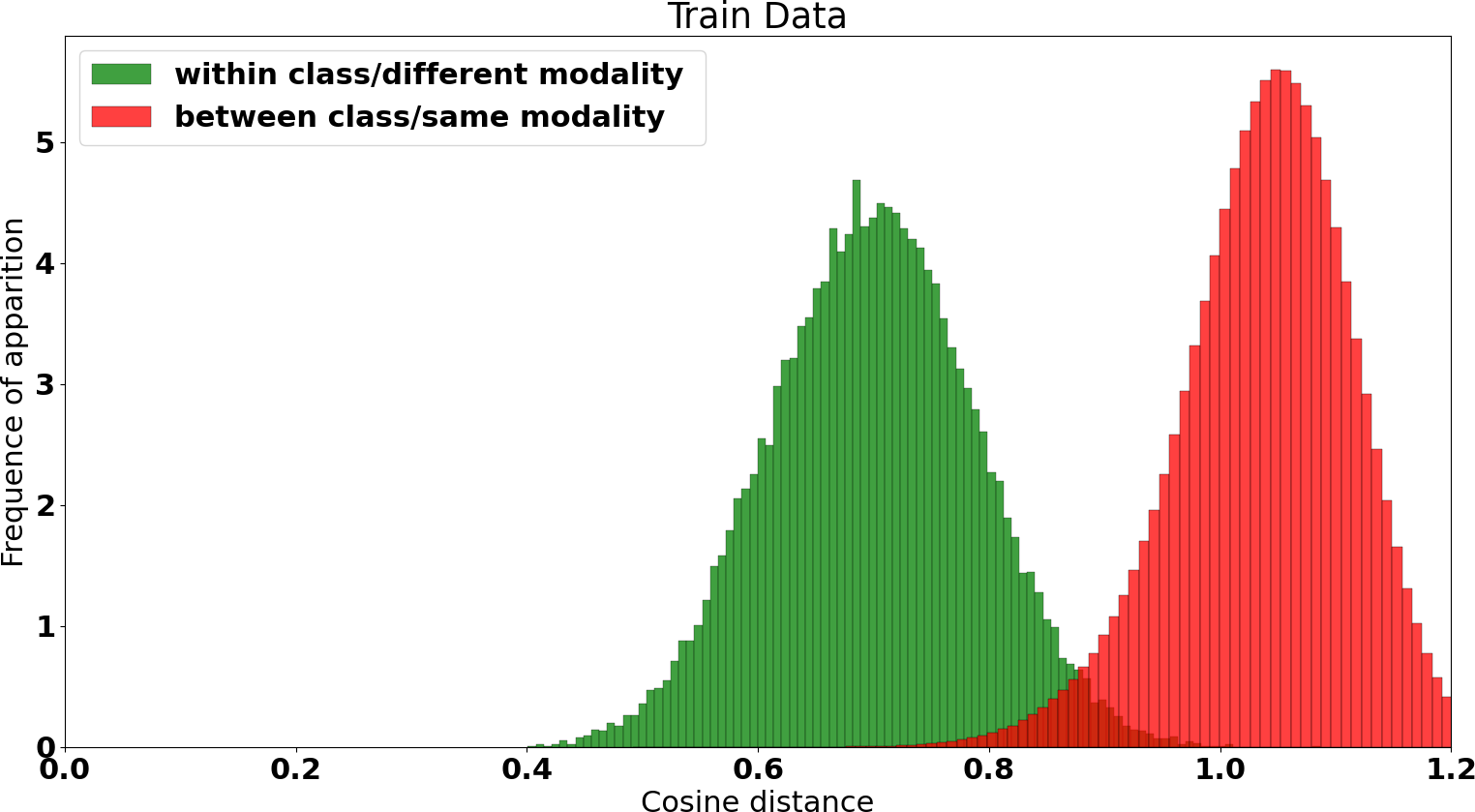}} & \raisebox{-.5\height}{\includegraphics[width=5.2cm]{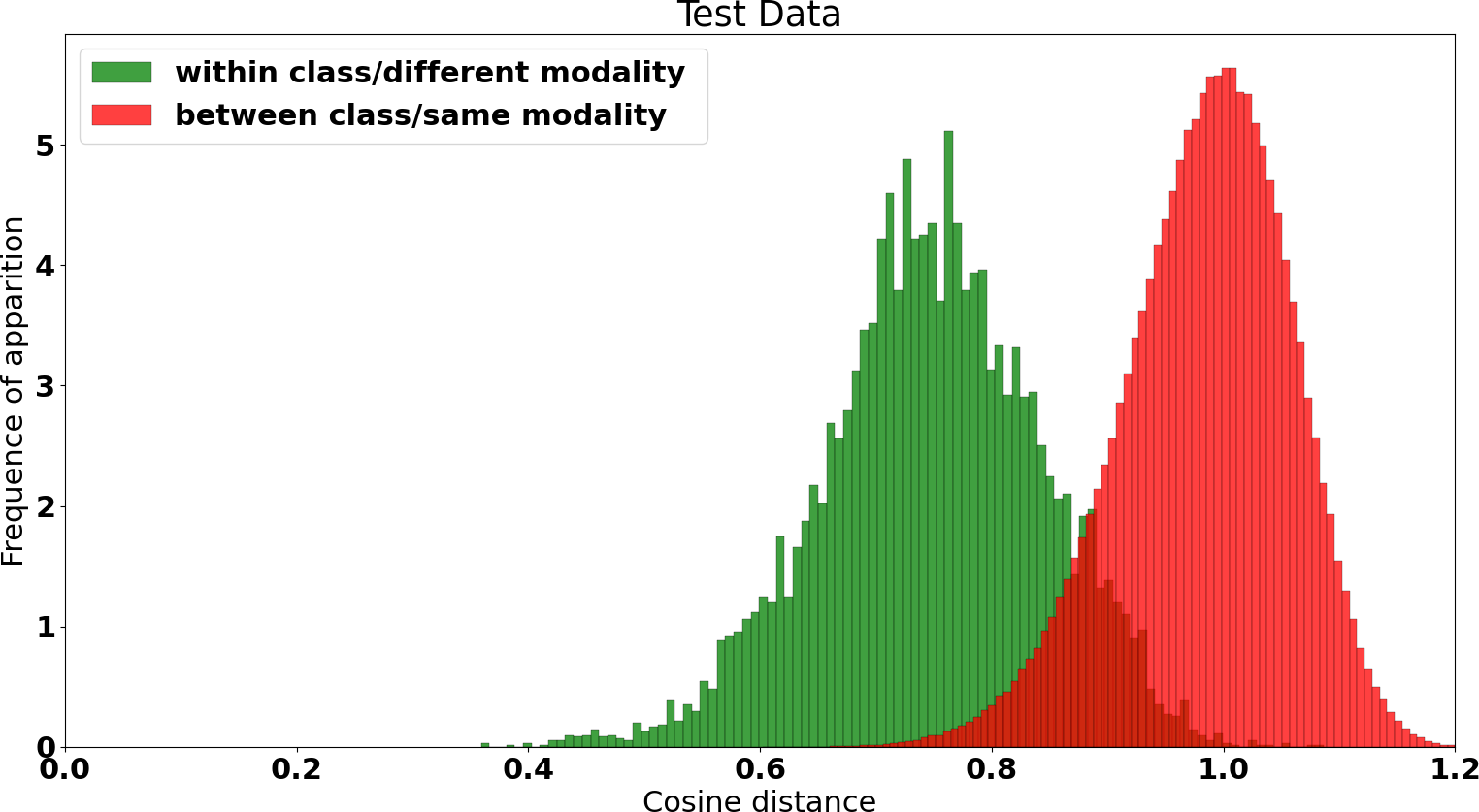}}    
\end{tblr}
}
\caption{Illustration domain shift between visible and infrared images. (a) Visual Images, (b) Infrared Images and (c) Select a random linear combination of RGB channels. }
\label{fig:q1}
\end{figure}

A visual representation of the cosine distance distributions for positive and negative cross-modality matching pairs has been provided in Fig. \ref{fig:q1} for both training and testing sets.  The proposed LUPI approaches are compared to the baseline (trained identity plus triplet loss, a strong baseline \cite{all-survey}) and the initial state. As the discrepancy between infrared and coloured visible images is enormous, the distance between the same identity in different modalities is greater than the distance between the same identity in different modalities. After training, such relationships should be reversed so that both strategies perform well in separating the infrared and RGB visible images in training. Despite this, the color-free intermediate method performed better than the baseline and gray intermediate in the testing set when incorporating the color-independent augmented one-channel images. By leveraging privileged intermediate information during training, our training approach improves the model's generalizability on the testing set to discriminate the visible and infrared images.

\begin{figure}[!h]
\centering
\centerline{
\begin{tblr}{
  colspec = {Q[m]X[c]},
}
(a)
& \raisebox{-.5\height}{\includegraphics[width=12cm]{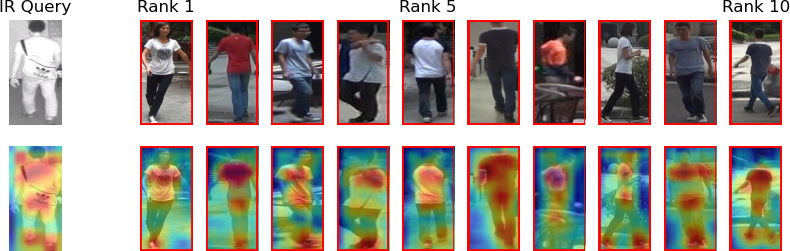}}  \\ 
(b) 
& \raisebox{-.5\height}{\includegraphics[width=12cm]{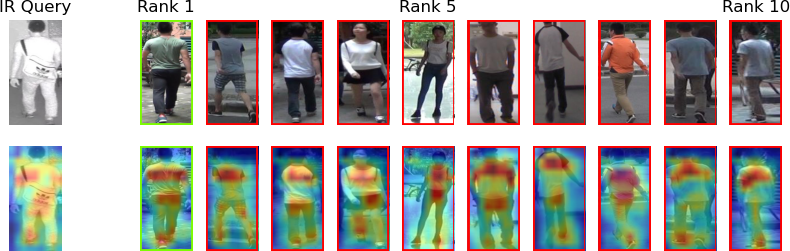}}      \\
(c) 
& \raisebox{-.5\height}{\includegraphics[width=12cm]{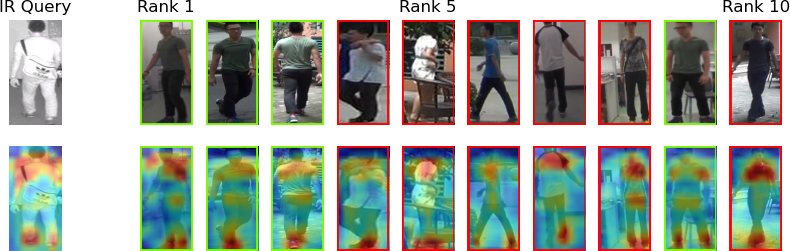}}
\end{tblr}
}
\caption{SYSU-MM01 Rank-10 visualizations for (a) Baseline (b) Gray and (c) Our, along with their corresponding similarity CAM \cite{stylianou2019visualizing}. Good and wrong matches are respectively surrounded by a green or a red box.}
\label{fig:vizualizations_rank_10}
\end{figure}

As another qualitative evaluation, the rank-10 for a given class of the SYSU-MM01 dataset is presented along with the corresponding similarity CAMs \cite{stylianou2019visualizing}. We produced those visualizations for the baseline Fig. \ref{fig:vizualizations_rank_10}.a, the gray model Fig. \ref{fig:vizualizations_rank_10}.b, and the random gray channel model Fig. \ref{fig:vizualizations_rank_10}.c. The query CAM is produced from the similarity with the best match. From the baseline to the gray and finally to our model, the attention gets refined, as it seems to focus on shoulders, hips and feet for the baseline while the hips attention gradually decrease for the grey and even almost fully disappear for our model. Interestingly, this behavior is consistent from one matching to another. In facts, we believe the hips focus by the baseline could come from the important color changes between  the lower and the upper body, which probably is well discriminant for the visible part of the model. However, regarding the infrared modality, this color marker is much less reliable and even probably appear as a source of confusion while practicing the cross-modal ReID. Thanks to the gray intermediate domain, or to the random channel selection which appear as even more powerful, this phenomena apparently diminishes.    

\begin{figure}[!h]
\centering
\centerline{
\begin{tblr}{
  colspec = {Q[m]X[c]},
}
(a)
& \raisebox{-.5\height}{\includegraphics[width=10cm]{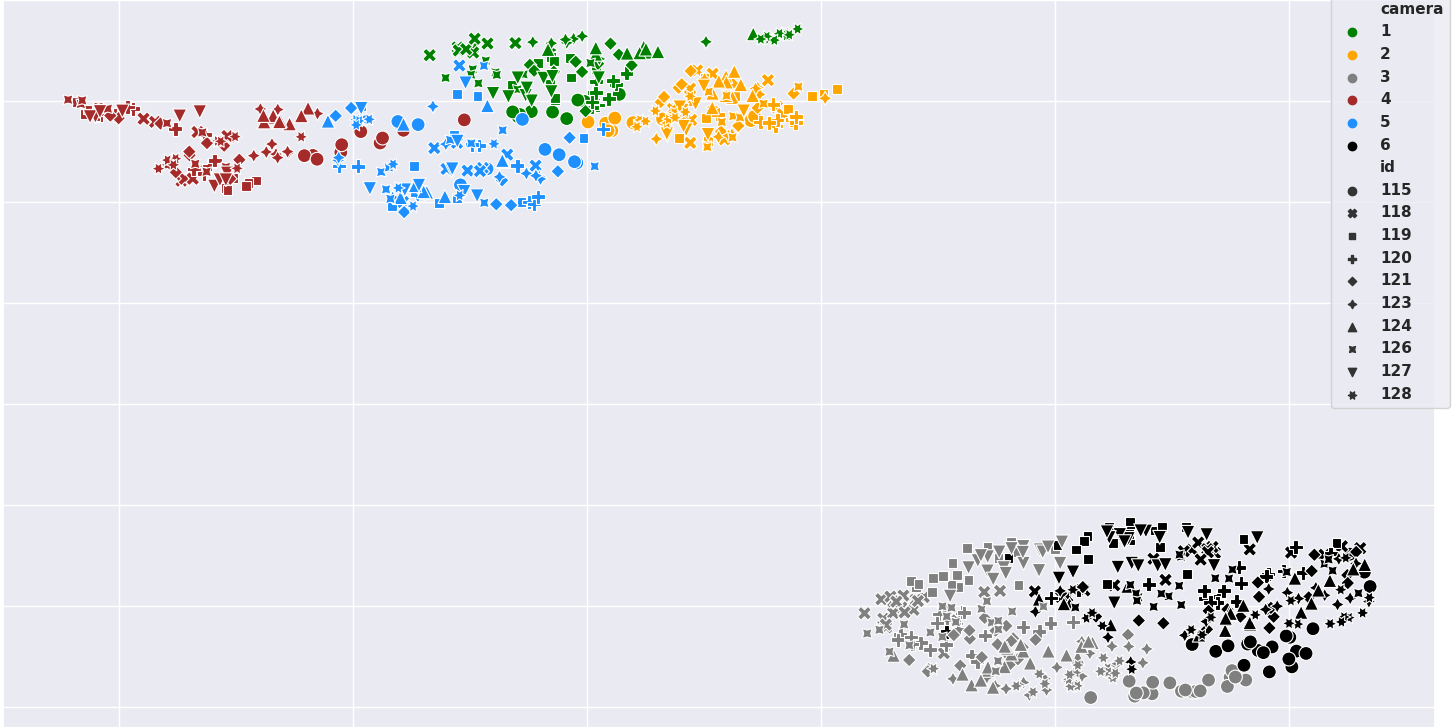}}  \\ 
(b) 
& \raisebox{-.5\height}{\includegraphics[width=10cm]{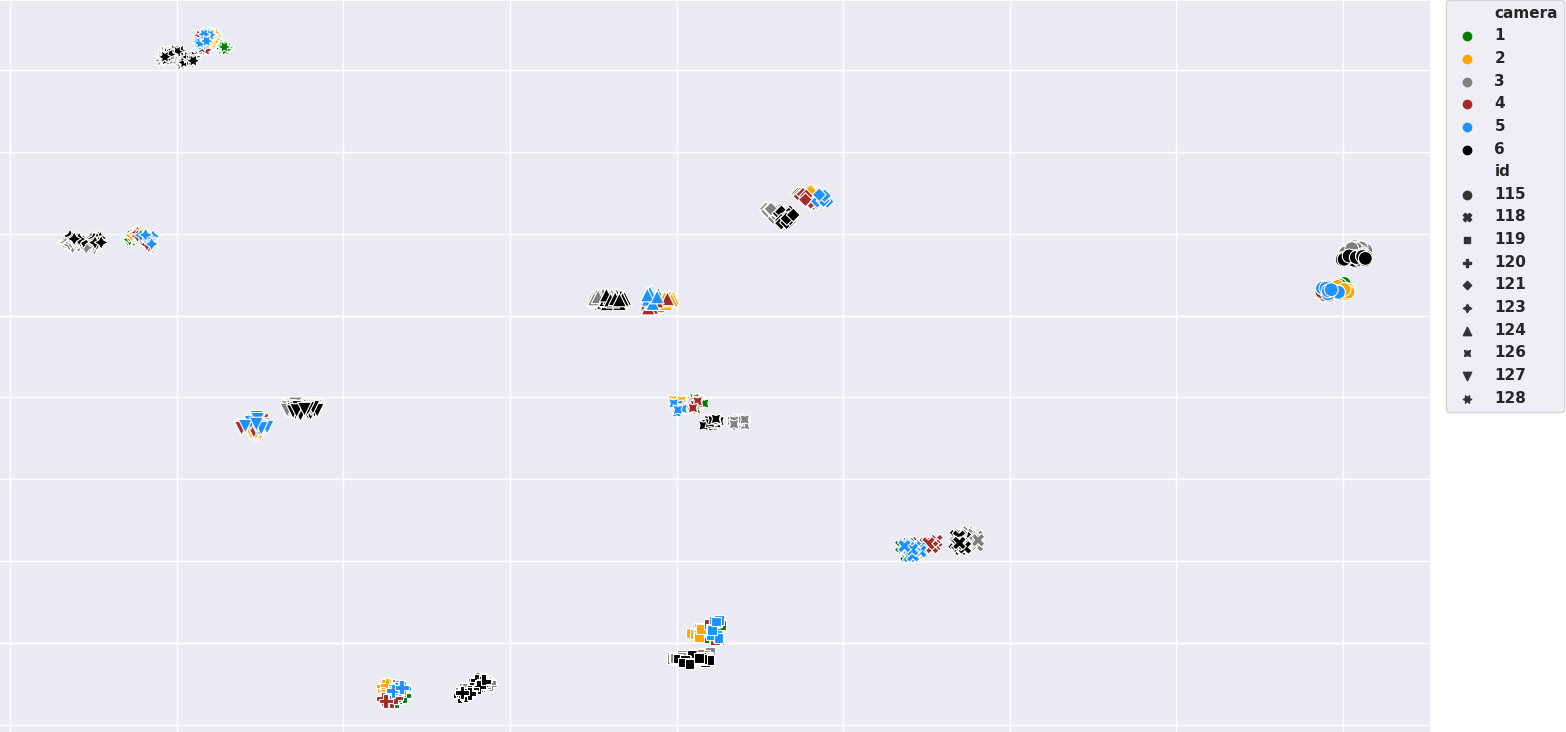}}      \\
(c) 
& \raisebox{-.5\height}{\includegraphics[width=10cm]{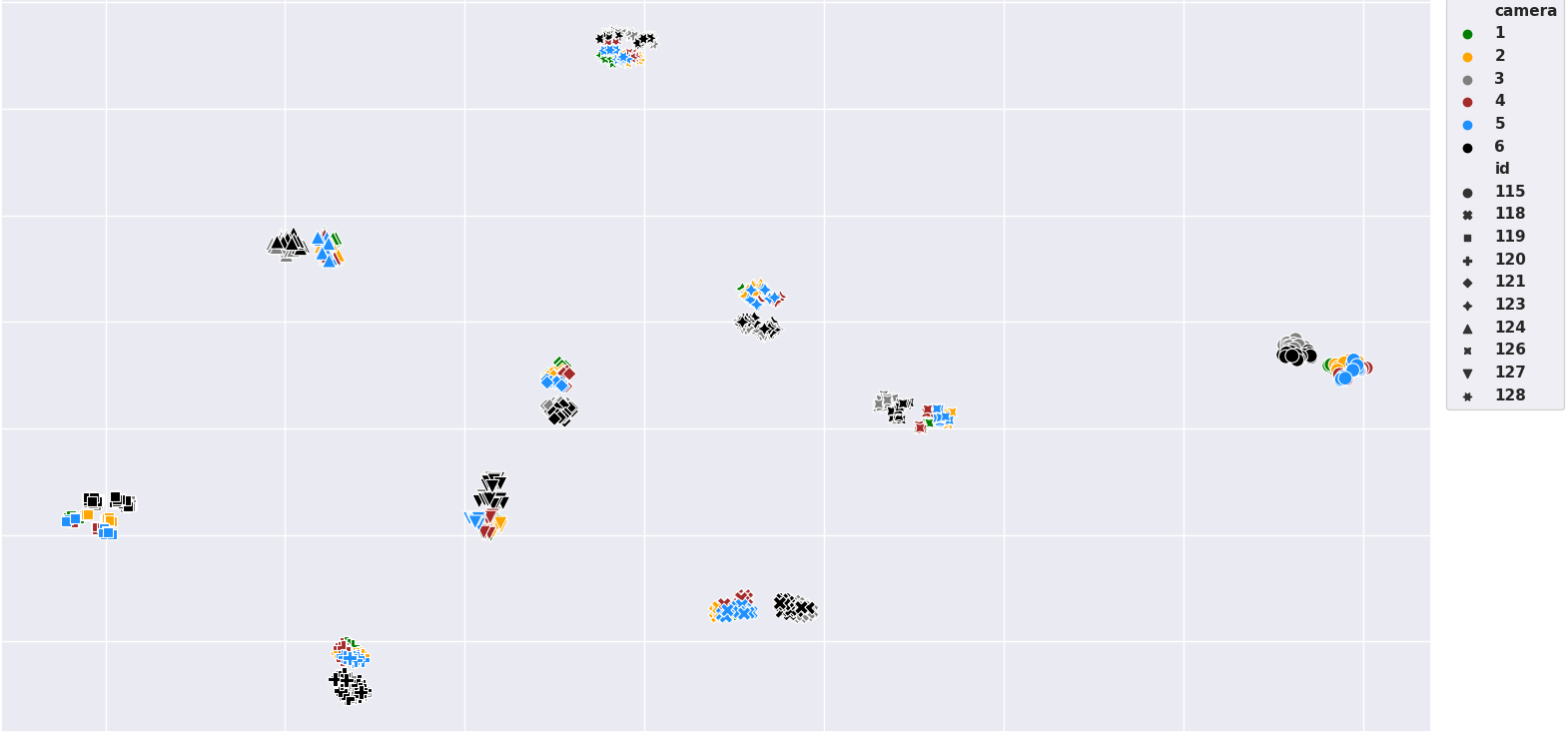}}     \\
(d) 
& \raisebox{-.5\height}{\includegraphics[width=10cm]{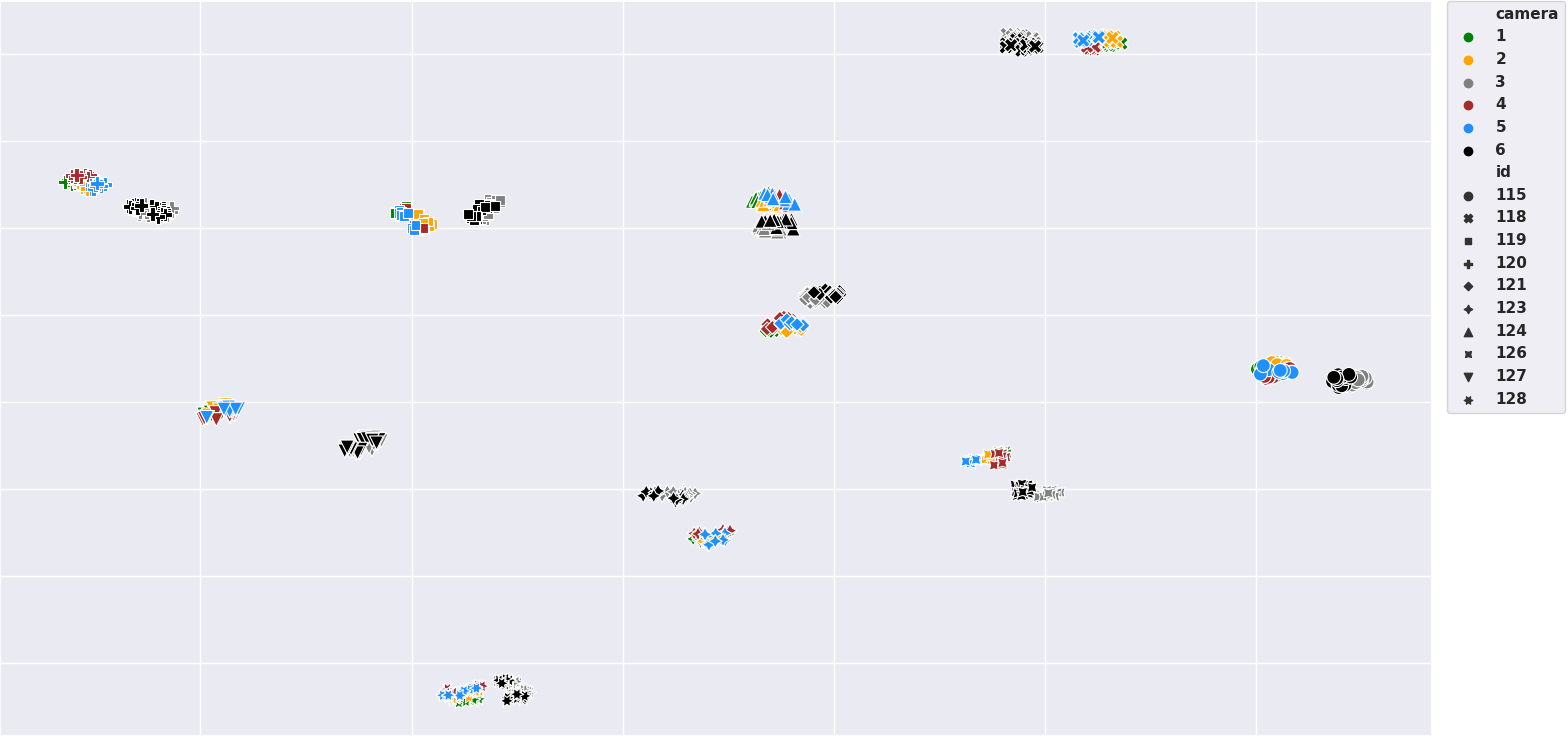}}     
\end{tblr}
}
\caption{UMAP \cite{mcinnes2018umap} visualization of ten randomly selected identities on the SYSUMM01 dataset from train sets. The feature extracted by (a) Initial, (b) Baseline (c) Gray and (d) Our models for 10 identities. Each identity is shown by the shape type and camera by the color. The gray and black colors mean the infrared camera.   }
\label{fig:umap}
\end{figure}

\begin{figure}[!h]
\centering
\centerline{
\begin{tblr}{
  colspec = {Q[m]X[c]},
}
(a)
& \raisebox{-.5\height}{\includegraphics[width=10cm]{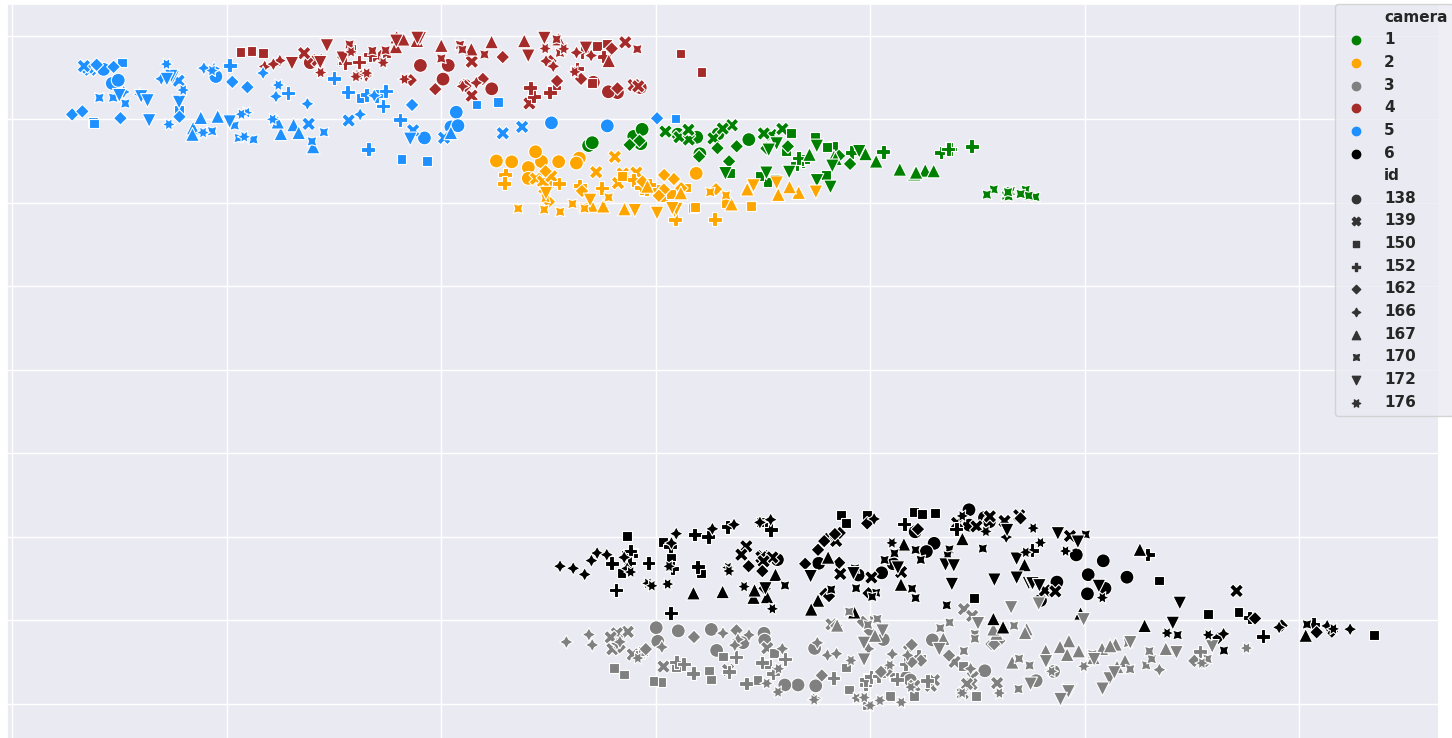}}  \\ 
(b) 
& \raisebox{-.5\height}{\includegraphics[width=10cm]{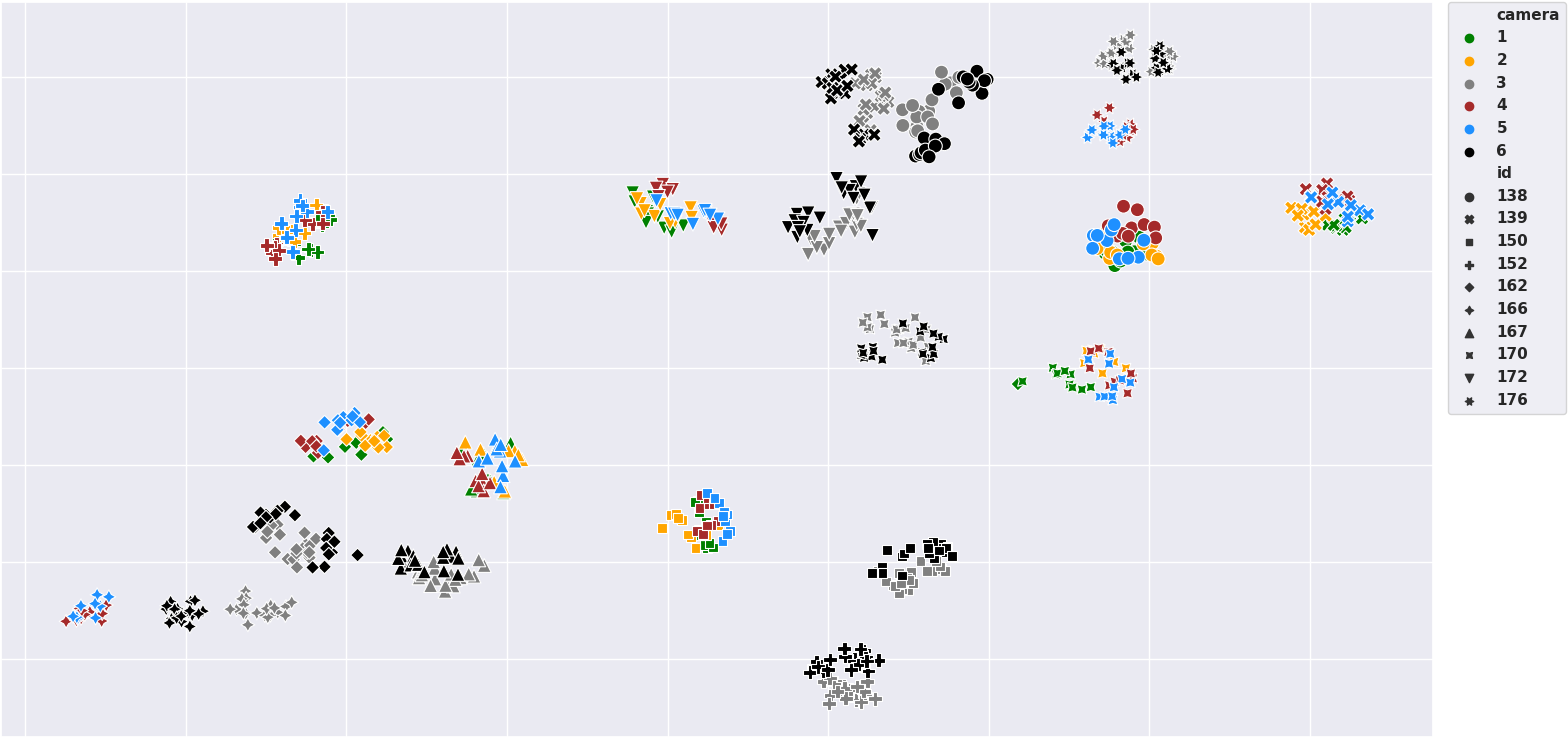}}     \\
(c) 
& \raisebox{-.5\height}{\includegraphics[width=10cm]{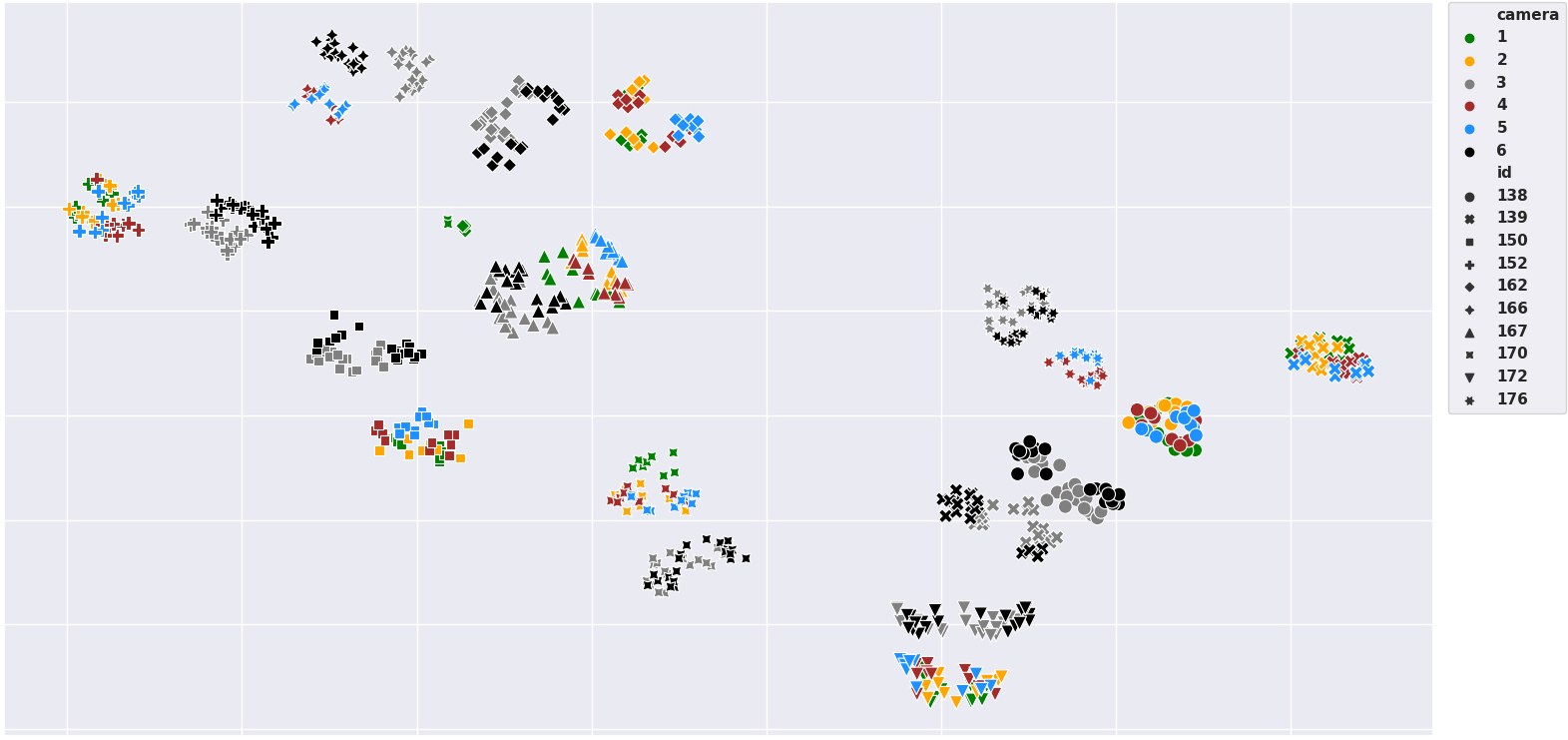}}    \\
(d) 
& \raisebox{-.5\height}{\includegraphics[width=10cm]{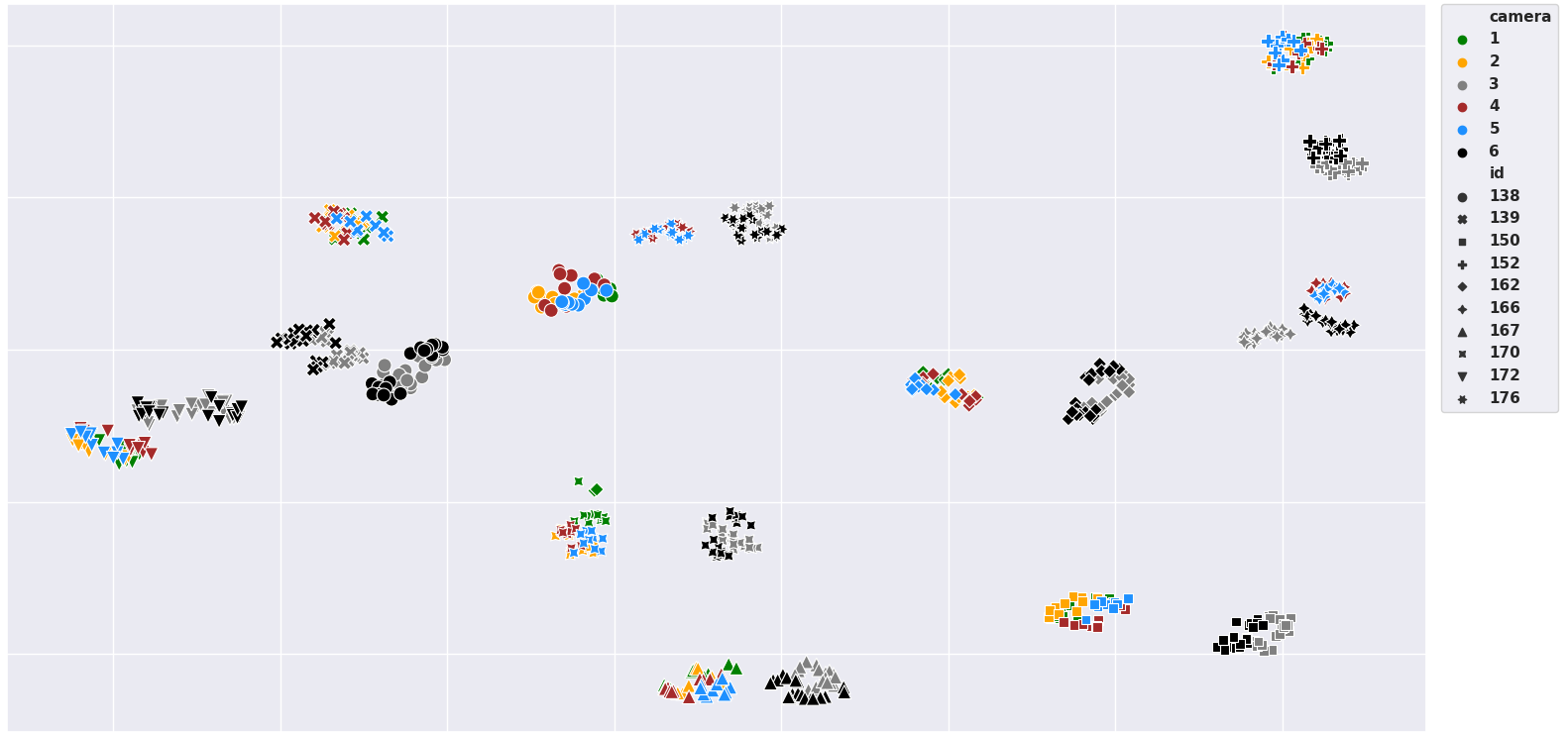}}    
\end{tblr}
}
\caption{UMAP \cite{mcinnes2018umap} visualization of ten randomly selected identities on the SYSUMM01 dataset from test. The feature extracted by (a) Initial, (b) Baseline (c) Gray and (d) Our models in two columns, first test and train identifies. Each identify is shown by the shape type and camera by the color. The gray and black colors mean the infrared camera.   }
\label{fig:umap_test}
\end{figure}

\clearpage
%
%
\bibliographystyle{splncs04}
\bibliography{paper}